\documentclass[final,a4paper]{article}
\usepackage{microtype}
\usepackage{graphicx}
\usepackage{subfigure}
\usepackage{booktabs} 
\usepackage{amsmath, amsfonts, bm, amsthm}
\usepackage{enumerate}
\usepackage{hyperref}
\usepackage{enumerate}
\usepackage{booktabs}

\newcommand{\x}{\mathbf{x}}
\newcommand{\y}{\mathbf{y}}

\theoremstyle{definition}
\newtheorem{theorem}{Theorem}[section]
\newtheorem{lemma}{Lemma}[section]
\newtheorem{proposition}{Proposition}[section]

\newtheorem*{definition}{Definition}

\usepackage{geometry}
\title{Learning Implicit Generative Models with Theoretical Guarantees}
\author{Yuan Gao\thanks{School of Mathematics and Statistics, Xi'an Jiaotong University, China (xjtuygao@gmail.com)}
\and Jian Huang\thanks{Department of Statistics and Actuarial Science, University of Iowa, Iowa City, IA 52242 (jian-huang@uiowa.edu)}
\and  Yuling Jiao\thanks{School of Mathematics and Statistics, Wuhan University, Wuhan 430072,  China. (yulingjiaomath@whu.edu.cn)}
\and Jin Liu\thanks{Center of Quantitative Medicine Duke-NUS Medical School, Singapore. (jin.liu@duke-nus.edu.sg)}
}
\begin{document}

\maketitle

\begin{abstract}
We propose a \textbf{uni}fied \textbf{f}ramework  for \textbf{i}mplicit \textbf{ge}nerative \textbf{m}odeling (UnifiGem) with theoretical guarantees by integrating approaches from optimal transport, numerical ODE,  density-ratio (density-difference) estimation and deep neural networks. First, the problem of implicit generative learning is formulated as that of finding the optimal transport map between the reference distribution and the target distribution, which is characterized by a totally nonlinear Monge-Amp\`{e}re equation. Interpreting the infinitesimal linearization of the Monge-Amp\`{e}re equation from the perspective of gradient flows in measure spaces leads to the continuity equation or the McKean-Vlasov equation. We then solve the McKean-Vlasov equation numerically using the forward Euler iteration, where the forward Euler map   depends  on the density ratio (density difference) between the  distribution at current iteration  and the underlying target distribution. We further estimate the density ratio (density difference) via deep density-ratio (density-difference) fitting and derive explicit upper bounds on the estimation error. Experimental results on both synthetic datasets and real benchmark datasets support our theoretical findings and demonstrate the effectiveness of UnifiGem.
\bigskip
\noindent
\textbf{Running title: UnifiGem}
\end{abstract}

\section{Introduction}
\label{intr}
The ability to efficiently model complex data and sample from complex distributions plays a key role in a variety of prediction and inference tasks in
machine learning and statistics \cite{salakhutdinov2015learning}.
The long-standing methodology for learning an underlying distribution relies on an explicit statistical data model, which can be difficult to specify in many modern machine learning tasks such as image analysis, computer vision and natural language processing.
In contrast, implicit generative models do not assume a specific form of the data distribution, but rather
learn a nonlinear map to transform a simple reference distribution to the underlying target distribution.   This modeling approach has been shown to achieve state-of-the-art performance in many machine learning tasks \cite{reed16,zhu17}.
\emph{Generative adversarial networks} (GAN) \cite{goodfellow14}, \emph{variational auto-encoders} (VAE) \cite{kingma14} and  \emph{flow-based methods} \cite{rezende2015variational} are important representatives of implicit generative models.

GANs model the
low-dimensional latent structure via deep nonlinear factors. They are trained by sequential differentiable surrogates of two-sample tests,  including the density-ratio test \cite{goodfellow14,nowozin16,mao17,mroueh17,tao18} and the density-difference test  \cite{li15,sutherland16,li17,arjovsky17,binkowski18} among others.
VAE is a probabilistic deep nonlinear factor model trained with  variational inference and stochastic approximation.
Several authors have proposed improved versions of VAE
by enhancing  the  disentangled representation power of the learned  latent codes  and reducing the blurriness of the generated images in vanilla VAE
 \cite{makhzani15,higgins16,tolstikhin18,zhang2019wasserstein}.
Flow-based methods learn a diffeomorphism map between the reference distribution and the target  distribution by maximum likelihood using the change of variables formula.
Recent work on flow-based methods has been focused on developing training methods and designing neural network architectures to trade off between the efficiency of training and sampling and the representation power of the learned map
\cite{rezende2015variational,dinh2014nice,dinh2016density,
kingma2016improved,
papamakarios2017masked,kingma2018glow,grathwohl2018ffjord}.

In this paper, we propose a unified framework  (UnifiGem)  for implicitly learning an underlying generative model by integrating approaches from optimal transport, numerical ODE,  density-ratio (density-difference) estimation and deep neural networks. The key idea  of implicit generative learning is to find a nonlinear  transform that pushes forward a simple reference distribution to the target distribution.  Mathematically, this task is known as finding an optimal transport map characterized by the Monge-Amp\`{e}re equation. However, it is quite challenging to solve the Monge-Amp\`{e}re equation due to the nonlinearity and high-dimensionality even at the population level assuming the target distribution is known.
The infinitesimal linearization of the Monge-Amp\`{e}re equation can be interpreted from the perspective of gradient flows in measure spaces,
which leads to the continuity equation.
Therefore, we turn to solve the continuity equation or equivalently, the characteristic ODE system associated with the continuity equation, which is a kind of McKean-Vlasov equation.
We solve the resulting McKean-Vlasov equation numerically using the forward Euler method and bound the discretization error at the population level since the forward Euler map depends on  the  density ratio (density difference) between the  distribution at current iteration and the underlying target distribution. We estimate the density ratio (density difference) via nonparametric deep density-ratio (density-difference) fitting and derive an explicit estimation error bound. Experimental results on both synthetic datasets and real benchmark datasets support our theoretical findings and demonstrate the effectiveness of UnifiGem.
\section{Notation, background and theory}
\label{back}
Let $\mathcal{P}_2(\mathbb{R}^{m})$ denote the space of Borel probability measures on $\mathbb{R}^{m}$ with finite second moments, and let $\mathcal{P}_2^{a}(\mathbb{R}^{m})$ denote the subset of $\mathcal{P}_2(\mathbb{R}^{m})$ in which measures are absolutely continuous with respect to the Lebesgue measure (all distributions are assumed to satisfy this assumption hereinafter).
$\mathrm{Tan}_{\mu}\mathcal{P}_2(\mathbb{R}^{m})$ denotes the tangent space  to $ \mathcal{P}_2(\mathbb{R}^{m})$ at $\mu$. Let  $\mathrm{AC}_{\mathrm{loc}}(\mathbb{R}^+,\mathcal{P}_2(\mathbb{R}^{m})) := \{\mu_t : I \rightarrow \mathcal{P}_2(\mathbb{R}^{m}) \ \  \mathrm{is \ \ absolutely \ \ continuous},  \ \  |\mu_t^{\prime}| \in L^{2}(I), I \subset \mathbb{R}^+\}$. $\mathrm{Lip}_{\mathrm{loc}}(\mathbb{R}^{m})$ denotes the set of functions that are Lipschitz continuous on any compact set of $\mathbb{R}^{m}$.
For any $\ell \in [1,\infty],$ we use  $L^{\ell}(\mu, \mathbb{R}^{m})$  ($L^{\ell}_{\mathrm{loc}}(\mu, \mathbb{R}^{m})$) to denote the
$L^{\ell}$ space of $\mu$-measurable functions on $\mathbb{R}^{m}$ (on any compact set of $\mathbb{R}^{m}$).
With $\textbf{I}$,  $\mathrm{det}$ and $\mathrm{tr}$, we refer to the identity map,  the determinant and the trace. We use $\nabla$, $\nabla^2$ and $\Delta$ to denote the gradient or Jacobian operator, the Hessian operator and the Laplace operator, respectively.

We first describe the theoretical background
used  in deriving UnifiGem, a unified framework to learn the  generative model $\nu$  implicitly from an i.i.d. sample $\{{X}_i\}_{i=1}^n \subset \mathbb{R}^{m}$.
\subsection{Wasserstein distance and optimal transport}
The quadratic Wasserstein distance between $\mu$ and $\nu \in \mathcal{P}_2(\mathbb{R}^{m})$  is defined as \cite{villani2008optimal,ambrosio2008gradient}
\begin{align} \label{wd}
\mathcal{W}_{2}(\mu, \nu) = \{ \inf_{\gamma \in \Gamma(\mu, \nu)} \mathbb{E}_{ (X, Y) \sim \gamma} [ \Vert X - Y \Vert_2^2 ] \}^{\frac12},
\end{align}
where $\Gamma(\mu, \nu)$ denotes the set of  couplings of $(\mu, \nu)$.
The static formulation of  $\mathcal{W}_{2}$ in  (\ref{wd}) admits the following  variational form   \cite{benamou2000computational}
\begin{align*}
  \mathcal{W}_2(\mu, \nu) &= \{ \inf_{q_t, \mathbf{v}_t} \{ \int_{0}^{1} \mathbb{E}_{\mathbf{X} \sim q_t} [ \Vert \mathbf{v}_t (\textbf{X}) \Vert_2^2 ] d t \} \}^{\frac12}, \\
\mathrm{s}.\mathrm{t}.\  \partial_t q_t(\x) &= - \nabla \cdot \left( q_t(\x) \mathbf{v}_t(\x) \right), \\
                  q_0(\x) &= q(\x), q_1(\x) = p(\x),
\end{align*}
where $\textbf{v}_t(\x): \mathbb{R}^+ \times \mathbb{R}^{m}  \rightarrow \mathbb{R}^{m}$ is a velocity vector field.
The Wasserstein distance $\mathcal{W}_{2}(\mu, \nu)$  measures the optimal quadratic  cost of  transporting  $\mu$ onto $\nu$. The corresponding optimal transport map $\mathcal{T}$ such that $\mathcal{T}_{\#} \mu = \nu$  is characterized by the  Monge-Amp\`{e}re equation \cite{brenier1991polar,mccann1995existence,santambrogio2015optimal}.
\begin{lemma}\label{lem1}
Let $\mu$ and $ \nu \in \mathcal{P}_2^{a}(\mathbb{R}^{m})$ with densities $q$ and $p$ respectively.  Then \eqref{wd} admits a unique solution $\gamma = (\textbf{I},\mathcal{T})_{\#} \mu$ with $$\mathcal{T} = \nabla \Psi,  \mu \text{-} a.e.,$$  where the potential function  $\Psi$ is convex and satisfies the Monge-Amp\`{e}re equation
\begin{equation}\label{mae}
\mathrm{det}(\nabla^2 \Psi(\x))= \frac{q(\x)}{p(\nabla \Psi(\x))}, \
 \x\in \mathbb{R}^{m}.
\end{equation}
\end{lemma}
It is challenging to find the optimal transport map $\mathcal{T}$ by  solving the totally nonlinear degenerate elliptic Monge-Amp\`{e}re equation \eqref{mae}. Linearization via a residual type of pushforward map, i.e.,  letting \begin{equation}\label{rm}
   \mathcal{T}_{t,\Phi} = \nabla \Psi = \textbf{I}  + t \nabla \Phi
    \end{equation}
    with a specially designed function $\Phi: \mathbb{R}^{m}  \rightarrow \mathbb{R}^1$ and a small $t\in \mathbb{R}^+$,
is a commonly used technique to address the difficulty due to nonlinearity
\cite{villani2008optimal}.
To be precise, let $X \sim q$, $\widetilde{X} = \mathcal{T}_{t,\Phi}(X),$ and
denote the distribution of $\widetilde{X}$ as $\widetilde{q}.$
With a small $t$, the map $\mathcal{T}_{t,\phi} $ is invertible according to the implicit function theorem, and we have the change of variables formula
\begin{equation}\label{cv}
 \mathrm{det}(\nabla^2 \Psi)(\x) =  |\mathrm{det}(\nabla \mathcal{T}_{t,\Phi})(\x)| = \frac{q(\x)}{\widetilde{q}(\tilde{\x})},
 \end{equation}
where
\begin{equation}\label{xc}
\tilde{\x} = \mathcal{T}_{t,\Phi}(\x).
\end{equation}
Using the fact
$\left.\frac{\mathrm{d}}{\mathrm{d} t}\right|_{t=0} \mathrm{det}(\textbf{A}+t \textbf{B})=\mathrm{det}(\textbf{A}) \mathrm{tr}\left(\textbf{A}^{-1} \textbf{B}\right)$  $\forall \textbf{A}, \textbf{B} \in \mathbb{R}^{m \times m}$ with $\textbf{A}$ invertible, and applying the first order Taylor expansion to \eqref{cv}  we have
\begin{align}
\log \widetilde{q}(\tilde{\x}) - \log q(\x)
= - t \Delta \Phi(\x) + o(t). \label{dc}
\end{align}
Let $t\rightarrow 0$ in \eqref{xc} and \eqref{dc},  we obtain
 a random process $\{\x_t\}$ and its law $q_t$ satisfying
\begin{align}
\frac{\mathrm{d} \x_t}{\mathrm{d} t} &= \nabla \Phi(\x_t), \ \  \mathrm{with} \ \  \x_0  \sim q, \label{leq1}\\
\frac{\mathrm{d} \ln  q_t(\x_t)}{\mathrm{d} t} &= - \Delta 
\Phi (\x_t), \ \  \mathrm{with} \ \  q_0 = q.\label{leq2}
\end{align}
Equations \eqref{leq1} and \eqref{leq2} resulting from  linearization  of the Monge-Amp\`{e}re equation \eqref{mae}
 can be interpreted as
 gradient flows in measure spaces \cite{ambrosio2008gradient}.
 And thanks to this connection, we can resort to solving a continuity equation characterized by a type of McKean-Vlasov equation, an ODE system that is easier to handle.
\subsection{Gradient flows in $ \mathcal{P}_2^a (\mathbb{R}^{m})$  }
For $\mu \in \mathcal{P}_2^a (\mathbb{R}^{m})$ with density $q$, let
\begin{equation}
\label{energyFun}
\mathcal{L}[\mu] = \int_{\mathbb{R}^{m}} F(q(\x)) {\rm d} \x: \mathcal{P}_2^a (\mathbb{R}^{m})  \rightarrow \mathbb{R}^{+} \cup \{0\}
\end{equation}
be an energy functional satisfying $\nu \in \arg\min \mathcal{L}[\cdot] ,$ where $F(\cdot): \mathbb{R}^{+} \rightarrow \mathbb{R}^{1}$ is a twice-differentiable convex function.
Among the widely used metrics on $ \mathcal{P}_2^a (\mathbb{R}^{m})$ in implicit generative learning, the following   two   are important examples of $\mathcal{L}[\cdot].$
\begin{itemize}
\item  $f$-divergence \cite{ali1966general}:
\begin{equation}\label{fdiv}
\mathbb{D}_f(\mu \Vert \nu) = \int_{\mathbb{R}^{m}} p(\x) f\left(\frac{q(\x)}{p(\x)} \right) {\mathrm{d}} \x,
\end{equation}
where $f: \mathbb{R}^+ \rightarrow \mathbb{R} $ is a twice-differentiable convex 
function satisfying $f(1) = 0$.
\item Lebesgue norm of density difference:
\begin{equation}\label{lm}
\|\mu-\nu\|^2_{L^2(\mathbb{R}^{m})} =  \int_{\mathbb{R}^{m}} |q(\x)- p(\x)|^2  {\mathrm{d}} \x.
\end{equation}
\end{itemize}

\begin{definition}
We call   $\{\mu_t\}_{t\in \mathbb{R}^+} \subset \mathrm{AC}_{\mathrm{loc}}(\mathbb{R}^+,\mathcal{P}_2(\mathbb{R}^{m}))$  a  gradient flow of the  functional $\mathcal{L}[\cdot]$,   if
 $\{\mu_t\}_{t\in \mathbb{R}^+} \subset  \mathcal{P}_2^a (\mathbb{R}^{m})$ $a.e.,   \ t \in \mathbb{R}^{+}$
and  the  velocity vector field $\textbf{v}_t \in \mathrm{Tan}_{\mu_t}\mathcal{P}_2 (\mathbb{R}^{m})$ satisfies
  $$\textbf{v}_t \in -\partial \mathcal{L}[\mu_t] \quad a.e. \quad t \in \mathbb{R}^+,$$
  where $\partial \mathcal{L}[\cdot]$ is the subdifferential of $\mathcal{L}[\cdot]$.
\end{definition}

The gradient flow $\{\mu_t\}_{t\in \mathbb{R}^+}$ of $\mathcal{L}[\cdot]$  enjoys the following nice properties.
\begin{theorem}\label{th1}
\begin{enumerate}[(i)]
 \item The continuity equation
\begin{equation}\label{vfp}
\frac{\mathrm{d}}{\mathrm{d} t}\mu_t
 = -\nabla\cdot(\mu_t\textbf{v}_t)\ \ {\rm in} \ \ \mathbb{R}^+ \times \mathbb{R}^{m} \ \ \mathrm{with} \ \ \mu_0 = \mu,
\end{equation}
holds in the sense of distributions.
\item Representation of the velocity fields.\\
If the density $q_t$ of $\mu_t$ is differentiable, then
\begin{equation}\label{vr}
 \textbf{v}_t(\x) =  -\nabla F^{\prime}(q_t(\x)) \ \ \mu_t\text{-}a.e. \ \  \x \in \mathbb{R}^{m}.
 \end{equation}
 \item Energy decay along the gradient flow.\\
 $$\frac{{\rm d} }{{\rm d} t} \mathcal{L}[\mu_t] = - \|\textbf{v}_t\|^2_{L^2(\mu_t,\mathbb{R}^{m})} \quad a.e. \quad t \in \mathbb{R}^+.$$
In addition, $$\mathcal{W}_2 (\mu_t,\nu) = \mathcal{O}(\exp^{- \lambda t}),$$ if $\mathcal{L}[\mu]$ is $\lambda$-geodetically convex with $\lambda>0$.
\item Conversely, if $\{\mu_t\}_t$ is the solution of continuity equation  \eqref{vfp} in (i) with $\textbf{v}_t(\x)$  specified by \eqref{vr} in (ii), then
$\{\mu_t\}_t$ is a gradient flow of $\mathcal{L}[\cdot]$.
\end{enumerate}
\end{theorem}

\begin{proposition}\label{prop1}
If we let $\Phi$ be time-dependent in  \eqref{leq1}-\eqref{leq2}, i.e., $\Phi_t$,  then the linearized  Monge-Amp\`{e}re equations \eqref{leq1}-\eqref{leq2} $\Leftrightarrow $ the continuity equation \eqref{vfp}
by taking  $\Phi_t(\x)  = -F^{\prime} (q_t(\x)).$
\end{proposition}

Theorem \ref{th1} and Proposition \ref{prop1} imply that $\{\mu_t\}_t$, the solution of the continuality equation  \eqref{vfp} with
$\textbf{v}_t =  -\nabla F^{\prime}(q_t(\x)),$
approximates the Monge-Amp\`{e}re equation \eqref{mae} and  converges rapidly  to the target distribution $\nu$.
Furthermore,   the continuity equation has the following representation under mild regularity conditions on the velocity fields.

\begin{theorem}\label{th2}
Assume $\|\textbf{v}_t\|_{L^{1}(\mu_t,\mathbb{R}^{m})}\in L^{1}_{\mathrm{loc}}(\mathbb{R}^+)$
and  ${\rm v}_t(\cdot)\in \mathrm{Lip}_{\mathrm{loc}}(\mathbb{R}^{m})$ with upper bound $B_t$ and Lipschitz constant $L_t$ such that $(B_t + L_t) \in  L^{1}_{\mathrm{loc}}(\mathbb{R}^+).$
Then the solution of  the continuity equation  \eqref{vfp} can be represented as
\begin{equation}\label{repc}
\mu_t = (\textbf{X}_t)_{\#}\mu,
\end{equation}
 where  $\textbf{X}_t(\x): \mathbb{R}^+ \times \mathbb{R}^{m} \rightarrow \mathbb{R}^{m} $ satisfies the  McKean-Vlasov equation
\begin{equation}\label{mve}
\frac{{\rm d}}{{\rm d} t} \textbf{X}_t(\x) = \textrm{v}_{t}(\textbf{X}_t(\x)) \ \ \mathrm{with} \ \  \textbf{X}_0  \sim \mu,
\end{equation}
$\mu$- a.e. $\x \in \mathbb{R}^{m}.$
\end{theorem}

We use the forward Euler method to solve the McKean-Vlasov equation \eqref{mve}.
Let   $s >0$ be a small step size. The forward Euler method is defined iteratively as follows:
\begin{align}
\mathcal{T}_{k} &= \textbf{I} + s \textbf{v}_{k},\label{eu1}\\
 \textbf{X}_{k+1}& = \mathcal{T}_{k}(\textbf{X}_k), \label{eu2}\\
\mu_{k+1} &=  (\mathcal{T}_{k})_{\#}  \mu_k, \label{eu3}
\end{align}
where $\textbf{X}_0 \sim \mu$, $\mu_0 = \mu$ and $k = 0,1,...,K$.
It is well known
that for a finite time horizon $T$ and a fixed compact domain, Euler discretization of the McKean-Vlasov equation \eqref{mve} has a global error of $\mathcal{O}(s)$ in the supremum norm \cite{leveque2007finite}.
Let $$\{\mu_t^{s}:  t\in [ks,(k+1)s)\}$$ be a piecewise linear  interpolation between $\mu_{k}$ and  $\mu_{k+1}.$
The  discretization error of  $\mu_t$ and  $\mu_{t}^{s}$ can be bounded in a finite time interval $[0,T)$.
\begin{proposition}\label{prop2}
$$\mathcal{W}_2(\mu_t, \mu_t^s) = \mathcal{O}(s).$$
\end{proposition}

Proposition  \ref{prop2} and (iii) in  Theorem \ref{th1} imply that the distribution of   the particles  $\textbf{X}_k$  defined in \eqref{eu2} with $k$ large enough is close to the target $\nu$.
The above theoretical  results are obtained at the population level, where  $\textbf{v}_k$  depends on the target $\nu$.
Therefore, it is natural  to implicitly
learn $\nu$  via first estimating  the discrete velocity fields $\textbf{v}_k$  at the sample level and then plugging the estimator of $\textbf{v}_k$ into \eqref{eu2}.
As shown  in  Lemma \ref{lem2} below, the velocity fields associated with the $f$-divergence \eqref{fdiv} and the Lebesgue norm \eqref{lm} are determined by density ratio and density difference respectively.
\begin{lemma}\label{lem2} The velocity fields $\mathbf{v}_t$ satisfy
\begin{equation*}
\mathbf{v}_{t}(\x) =
\left\{\begin{array}{ll}
-f^{\prime\prime}(r_t(\x))\nabla r_t(\x), \ \  \mathcal{L}[\mu] = \mathbb{D}_f(\mu \Vert \nu), \\
- 2\nabla d_t(\x),  \ \ \mathcal{L}[\mu] = \|\mu-\nu\|^2_{L^2(\mathbb{R}^{m})},
\end{array}
\right.
\end{equation*}
where   $$  r_t(\x) = \frac{q_t(\x)}{p(\x)} \ \text{ and }\
d_t(\x) = q_t(\x) -  p(\x), \x \in \mathbb{R}^m .$$
\end{lemma}
Several methods have been developed to estimate  density ratio and density difference in the literature. Examples include probabilistic classification approaches, moment matching and direct density-ratio (density-difference) fitting, see \cite{sugiyama2012density2,sugiyama2012density,kanamori2014statistical,mohamed2016learning} and the references therein.

\subsection{Deep density-ratio and density-difference fitting}
The evaluation of velocity fields depends on dynamic estimation of a discrepancy (density ratio or density difference) between the pushforward distribution $q_t$ and the target distribution $p$.
Density-ratio and density-difference fitting with the Bregman score provides a unified framework for such discrepancy estimation \cite{gneiting2007strictly,dawid2007geometry,sugiyama2012density2,sugiyama2012density,kanamori2014statistical} without estimating each probability distribution separately.

We use a neural network $R_\phi: \mathbb{R}^{m}\rightarrow \mathbb{R}^1$ with parameter $\phi$ to parameterize the density ratio $r(\x) = \frac{q(\x)}{p(\x)}$  between a given density  $q$ and the target $p$.   Let $g: \mathbb{R} \rightarrow \mathbb{R}$ be a differentiable and strictly convex function. The separable Bregman score with the base probability measure $p$ to measure the discrepancy between $R_\phi$ and $r$ is
\begin{align*}
&\mathfrak{B}_{\rm ratio}(r, R_{\phi}) \\
&= \mathbb{E}_{X \sim p} [ g^{\prime}(R_{\phi}(X)) (R_{\phi}(X) - r(X)) - g(R_{\phi}(X)) ] \\
&= \mathbb{E}_{X \sim p} [ g^{\prime}(R_{\phi}(X)) R_{\phi}(X) - g(R_{\phi}(X)) ] \\
&\ \ \ \ - \mathbb{E}_{X\sim q} [g^{\prime}(R_{\phi}(X))].
\end{align*}
And $\mathfrak{B}_{\rm ratio}(r, R_{\phi}) \ge \mathfrak{B}_{\rm ratio}(r, r)$, where the equality holds iff $R_{\phi} = r$.

For deep density-difference fitting, a neural network $D_{\psi}: \mathbb{R}^{m}\rightarrow \mathbb{R}^1$ with parameter $\psi$ is utilized to estimate the density-difference
$d(\x)= q(\x) - p(\x)$ between a given density $q$ and the target $p$. The separable Bregman score with the base probability measure $w$ to measure the discrepancy between $D_{\psi}$ and $d$ can be derived  similarly,
\begin{align*}
& \mathfrak{B}_{\rm diff}(d, D_{\psi}) \\
&= \mathbb{E}_{X \sim p} [w(X) g^{\prime}(D_{\psi}(X))] - \mathbb{E}_{X \sim q} [w(X) g^{\prime}(D_{\psi}(X))] \\
&\ \ \ \  + \mathbb{E}_{X\sim w} [ g^{\prime}(D_{\psi}(X)) D_{\psi}(X) - g(D_{\psi}(X)) ].
\end{align*}
Here, we focus on the widely used least-squares density-ratio (LSDR) fitting with $g(c) = (c-1)^2$ as a working example:
  \begin{align*}
\mathfrak{B}_{\rm LSDR}(r, R_{\phi})
= \mathbb{E}_{X \sim p} [ R_{\phi}(X) ^2 ]
- 2 \mathbb{E}_{X \sim q} [R_{\phi}(X)] + 1,
\end{align*}
  The scenario of other functions, such as $g(c) = c\log c - (c+1)\log(c+1)$ corresponding to estimating $r$ via the logistic regression (LR),  and the case of density-difference fitting can be handled similarly.

\subsection{Weighted Gradient penalties}
The distributions of real data may have a low-dimensional structure with their supports concentrated on a low-dimensional manifold, which   may cause the  $f$-divergence to be  ill-posed due to non-overlapping supports.
Motivated by recent works on smoothing via noise injection \cite{sonderby2016amortised,arjovsky2017principled} and Tikhonov regularization method for $f$-GAN \cite{roth2017stabilizing}, we derive a simple weighted gradient penalty  to improve deep density-ratio fitting.
We consider a noise convolution form of $\mathfrak{B}_{\rm ratio}(r, R_{\phi})$ with Gaussian noise  $\bm{\epsilon} \sim \mathcal{N}(\mathbf{0}, \alpha  \mathbf{I})$,
\begin{align*}
&\mathfrak{B}_{\rm ratio}^{\alpha}(r, R_{\phi}) \\
&= \mathbb{E}_{X\sim p} \mathbb{E}_{\bm{\epsilon} } [ g^{\prime}(R_{\phi}(X + \bm{\epsilon})) R_{\phi}(X + \bm{\epsilon}) - g(R_{\phi}(X + \bm{\epsilon})) ] \\
&\ \ \ \ - \mathbb{E}_{X \sim q} \mathbb{E}_{\bm{\epsilon} } [g^{\prime}(R_{\phi}(X + \bm{\epsilon}))].
\end{align*}
Taylor expansion applied to $R_{\phi}$ gives
\begin{align*}
\mathbb{E}_{\bm{\epsilon}} [R_{\phi}(\x + \bm{\epsilon})] = R_{\phi}(\x) + \frac{\alpha}{2} \Delta{R_{\phi}(\x)} + \mathcal{O}(\alpha^2).
\end{align*}
Using  equations (13)-(17) in \cite{roth2017stabilizing}, we get
\begin{align*}
\mathfrak{B}_{\rm ratio}^{\alpha}(r, R_{\phi})
\approx \mathfrak{B}_{\rm ratio}(r, R_{\phi}) + \frac{\alpha}{2} \mathbb{E}_{p} [g''(R_{\phi}) \Vert \nabla R_{\phi} \Vert_2^2 ],
\end{align*}
i.e., $\frac12 \mathbb{E}_{p} [g^{\prime\prime}(R_{\phi}) \Vert \nabla R_{\phi} \Vert_2^2 ]$ serves as a regularizer for deep
density-ratio fitting when $g$ is twice differentiable. As a consequence, for $g(c) = (c-1)^2$, the resulting gradient penalty
\begin{equation}\label{gp}
   \mathbb{E}_{p} [\Vert \nabla R_{\phi} \Vert_2^2],
\end{equation}
recovers the well-known squared Sobolev semi-norm in nonparametric statistics.
\subsection{Estimation error}
\begin{lemma}\label{lem3}
For given densities $p$ and $q$, let $r= \frac{q}{p}$ with
$ \mathcal{C} =  \mathbb{E}_{X\sim q} [r^2(X)] - 1 < \infty .$
For any $\alpha \geq 0$, define a nonnegative functional
$$\mathfrak{B}^{\alpha}_{\rm LSDR}(R) = \mathfrak{B}_{\rm LSDR}(r, R) + \alpha \mathbb{E}_{p} [\Vert \nabla R \Vert_2^2]+ \mathcal{C}.$$
Then,  $$r \in \arg\min_{\text{measureable}\, R } \mathfrak{B}^{0}_{\rm LSDR}(R).$$
And  $$\mathfrak{B}^{\alpha}(R)=0 \ \ \mathrm{iff} \ \  R(\x) = r(\x)  = 1  \ \ (q, p)\text{-}a.e. \  \x \in \mathbb{R}^m.$$
\end{lemma}

At the population level, according to Lemma \ref{lem3}, we can recover  the density ratio $r$  via minimizing $\mathfrak{B}^{\alpha}_{\rm LSDR}(R)$. Moreover, the gradient penalty \eqref{gp} stabilizes and improves the long time behavior of Euler iterations at the sample level, where the pushforward distribution should be close to the target as expected. This is supported by our numerical experiments in Section \ref{numerics}.

Let $\mathcal{H}_{\mathcal{D}, \mathcal{W}, \mathcal{S}, \mathcal{B}}$ be the set of ReLU  neural networks $R_{\phi}$ with  depth   $\mathcal{D}$,   width $\mathcal{W}$,  size $ \mathcal{S}$, and $ \|R_{\phi}\|_{\infty} \leq \mathcal{B}.$
At the sample level,
only i.i.d. data   $\{X_i\}_{i=1,...,n} $ and $\{Y_i\}_{i=1,...,n} $  sampled from $p$ and $q$ are available.
We estimate  $r$ with   $\widehat{R}_{\phi}$
defined as
\begin{equation}\label{sf}
\widehat{R}_{\phi} \in \arg \min_{R_\phi\in \mathcal{H}_{\mathcal{D}, \mathcal{W}, \mathcal{S}, \mathcal{B}}}  \sum_{i=1}^n   \frac{1}{n}(R_{\phi}(X_i)^2 +\alpha \|\nabla R_{\phi}(X_i)\|^2_{2}-2R_{\phi}(Y_i)).
\end{equation}
Next we bound the nonparametric estimation error  $\|\widehat{R}_{\phi} - r\|_{L^2(\nu)}$  under the assumption that the support of $\nu$   concentrates  on a  compact low-dimensional manifold and $r$ is Lipsichiz continuous.
Let $ \mathfrak{M } \subseteq[-c,c]^{m}$ be a
Riemannian manifold with dimension  $\mathfrak{m}$, condition number $1 / \tau$,  volume  $\mathcal{V}$,  geodesic covering regularity $\mathcal{R}$, and
$$\mathfrak{m}\ll  \mathcal{M} =
\mathcal{O}\left(\mathfrak{m} \ln (\mathfrak{m} \mathcal{V} \mathcal{R}/\tau)\right)
\ll m.$$
Denote
$\mathfrak{M}_{\epsilon} =\left\{\x \in[-c,c]^{m} : \inf \{\|\x-\y\|_2 : \y \in \mathfrak{M}\} \leq \epsilon\right\},$  $\epsilon \in(0,1)$.
\begin{theorem}\label{th3}
Assume  $\mathrm{supp}(r) = \mathfrak{M}_{\epsilon}$ and $r(\x)$ is Lipschitz continuous with the bound $B$ and the Lipschitz constant $L$.
Suppose the topological  parameter of  $\mathcal{H}_{\mathcal{D}, \mathcal{W}, \mathcal{S}, \mathcal{B}}$ in \eqref{sf} with $\alpha = 0$ satisfies $\mathcal{D} = \mathcal{O}(\log n)$, $ \mathcal{W} = \mathcal{O}(n^{\frac{\mathcal{M}}{2(2+\mathcal{M})}}/\log n)$,  $\mathcal{S} = \mathcal{O}(n^{\frac{\mathcal{M}-2}{\mathcal{M}+2}}/\log^4 n)$, and $\mathcal{B} = 2B$.
Then,
\begin{equation*}
\mathbb{E}_{\{X_i,Y_i\}_{1}^n} [\|\widehat{R}_{\phi} - r\|_{L^2(\nu)}^2] \leq C (B^2+ cL m \mathcal{M}) n^{-2/(2+ \mathcal{M})},
\end{equation*}
where $C$ is a universal constant.
\end{theorem}

\section{A unified framework for implicitly  deep generative modeling}\label{UnifiGem}
We are now ready to described how to implement UnifiGem with i.i.d.
 data $\{{X}_i\}_{i=1}^n \subset \mathbb{R}^{m}$  from an unknown target distribution $\nu$.
UnifiGem is a particle method, with which we learn a transport map that transforms particles from a simple reference distribution $\mu$, such as the standard normal distribution or the uniform distribution, into particles from the target distribution $\nu$. From Theorems \ref{th2} and \ref{th1} and Proposition \ref{prop2} we know that at the population level, the solution $\mathbf{X}_t$  of the McKean-Vlasov equation \eqref{mve} with a sufficiently large $t$ is a good approximation of such a transform. This solution can be obtained accurately via the forward Euler iteration  \eqref{eu1}-\eqref{eu3} with a small step size, i.e., $$ \mathcal{T}_K \circ\mathcal{T}_{K-1} \circ ... \circ\mathcal{T}_1$$ serves as a desired  transform with a large $K$.
As implied by Theorem \ref{th3}, each $\mathcal{T}_k, k = 1,...,K$ can be estimated with high accuracy by
$$\widehat{\mathcal{T}}_k = \mathbf{I} + s \hat{\mathbf{v}}_k$$
where $ \hat{\mathbf{v}}_k= -f^{\prime\prime}(\widehat{R}_{\phi})(\x))\nabla \widehat{R}_{\phi}(\x)$.  Here  $\widehat{R}_{\phi}$ is estimated based on \eqref{sf} with $\{Y_i\}_{i = 1,...n}\sim q_k$.
Therefore, the particles  $$\widehat{\mathcal{T}}_K \circ\widehat{\mathcal{T}}_{K-1} \circ ... \circ\widehat{\mathcal{T}}_1(\tilde{Y}_i), i = 1,...n$$ serve as samples drawn from the target distribution $\nu$, where particles $\{\tilde{Y}_i\}_{i=1}^n \subset \mathbb{R}^{m}$ are sampled from a simple reference distribution $\mu$.

In many applications, high-dimensional complex data such as images, texts and natural languages, tend to have low-dimensional features.
To learn generative models with hidden low-dimensional structures, it is beneficial to have the option of first sampling particles $\{Z_i\}_{i=1}^n$ from
a low-dimensional reference distribution $\tilde{\mu} \in \mathcal{P}_2(\mathbb{R}^{\ell})$ with $\ell \ll d$.
Then we apply  $$\widehat{\mathcal{T}}_K \circ\widehat{\mathcal{T}}_{K-1} \circ ... \circ\widehat{\mathcal{T}}_1$$
to particles $$\tilde{Y}_i = G_{\theta}(Z_i), i = 1,...n$$, where we introduce another deep neural network $G_{\theta}:\mathbb{R}^{\ell} \rightarrow \mathbb{R}^{m}$ with parameter $\theta$. We can estimate  $G_{\theta}$ via fitting  the pairs $\{(Z_i, \tilde{Y}_i)\}_{i=1}^n$.   We give a detailed description of the UnifiGem  algorithm below.
\begin{itemize}
\item \textbf{Outer loop for  modeling low dimensional latent structure (optional)}
	\begin{itemize}
	\item  Sample  $\{Z_i\}_{i=1}^n\subset \mathbb{R}^{\ell}$  from a low-dimensional simple reference distribution $\tilde{\mu}$ and let  $\tilde{Y}_i = G_{\theta}(Z_i), i =1,2,...,n$.
	\item \textbf{Inner loop for finding the pushforward map}
		\begin{itemize}
        \item If there are no outer loops, sample $\tilde{Y}_i \sim \mu, i =1,2,...,n$.
		\item   Get $\hat{\mathbf{v}}(\x) = -f^{\prime\prime}(\widehat{R}_{{\phi}}(\x))\nabla \widehat{R}_{{\phi}}(\x)$   via solving \eqref{sf} with $Y_i = \tilde{Y}_i$.
       Set $\widehat{\mathcal{T}} = \mathbf{I} + s \hat{\mathbf{v}}$ with small step size $s$.
		\item  Update the particles  $\tilde{Y}_i = \widehat{\mathcal{T}}(\tilde{Y}_i)$,   $i =1,2,...,n$.
		\end{itemize}
	\item \textbf{End inner loop}
	\item  If there are outer loops,
update the parameter $\theta$ of $G_\theta(\cdot)$ via solving   $\min_{\theta}\sum_{i=1}^n \|G_{\theta}(Z_i) - \tilde{Y}_i\|_{2}^2/n$.
    \end{itemize}
\item \textbf{End outer loop}
\end{itemize}
UnifiGem is a unified and general framework, since it allows different choices of the  energy functionals $\mathcal{L}[\cdot]$ in \eqref{energyFun}
and density-ratio (density-difference) estimators.
\section{Related works}\label{relate}
We discuss connections between UnifiGem and the existing related works, especially those that use optimal transport based on Wasserstein  distances and gradient flows in measure spaces.
Implicit generative learning aims at
finding a transform map that pushes forward a simple reference distribution $\mu$ to the target $\nu$. The existing implicit generative models, such as VAEs, GANs and flow-based methods, parameterize such a map with a neural network, say $G_{\theta}$, that solves
 \begin{equation}\label{ul}
 \min_{\theta} \mathfrak{D}((G_{\theta})_{\#} \mu,\nu),
 \end{equation} where $\mathfrak{D}(\cdot, \cdot)$ is an integral probability discrepancy.
$f$-GAN \cite{nowozin16} including the vanilla GAN \cite{goodfellow14},  and WGAN \cite{arjovsky17} solve the dual form of  \eqref{ul} via parameterizing the dual variable with another neural network with  $\mathfrak{D}$ as  the $f$-divergence and the $1$-Wasserstein distance respectively.
Based on the fact that the $1$-Wasserstein distance can be evaluated from  samples via linear programming \cite{srip12},
\cite{liu2018two} and  \cite{genevay2018learning} proposed training the primal form of WGAN  via a two-stage method that solves the linear programm  and
refits the optimal pairs with a neural network and unrolling the Sinkhorn iteration respectively. SWGAN \cite{deshpande2018generative} and
 MMDGAN \cite{li17,binkowski18} use the sliced quadratic Wasserstein  distance and the maximum mean discrepancy  (MMD) as the discrepancy $\mathfrak{D}$ respectively.

Vanilla VAE \cite{kingma14}  approximately solves the primal form of \eqref{ul} with the KL-divergence loss  under the framework of  variational inference.
Several authors have proposed methods that use optimal transport losses, such as various forms of Wasserstein  distances
between the distribution of learned latent codes and the prior distribution as the regularizer in VAE to improve performance. These methods include  WAE    \cite{tolstikhin18},  Sliced WAE \cite{modelsliced} and Sinkhorn AE \cite{patrinisinkhorn}.

Discrete time  flow-based methods \cite{rezende2015variational,dinh2014nice,dinh2016density,kingma2016improved,papamakarios2017masked,kingma2018glow} minimize \eqref{ul} with the KL divergence loss. \cite{grathwohl2018ffjord} proposed  an ODE flow   for fast training via using the adjoint  equation \cite{chen2018neural}.
By introducing  the optimal transport tools into maximum  likelihood training,  \cite{chen2018continuous} and \cite{zhang2018monge}    considered continuous time flow. \cite{chen2018continuous} proposed a gradient flow in measure spaces in the framework of variational inference  and  then discretized it with the implicit movement minimizing scheme  \cite{de1993new,jordan1998variational}.
  \cite{zhang2018monge} actually considered  gradient flows in measure spaces with time invariant velocity  fields.
CFGGAN \cite{johnson18} derived from the perspective of optimization in the functional space is exactly a special form of UnifiGem with $\mathcal{L}[\cdot]$ taken as the KL divergence.
 SW flow  \cite{liutkus2019sliced} and MMD flow   \cite{arbel2019maximum} are gradient flows in measure spaces.
These methods are most related to our proposed UnifiGem.
 In SW flow, the energy functional $\mathcal{L}[\cdot]$ in \eqref{energyFun} is the sliced quadratic Wasserstein distance  penalized  with the entropy regularizer. We should mention that with SW flow,  the target $\nu$ may not be the minimizer of such $\mathcal{L}[\cdot]$ even at the population level.
MMD flow can be recovered from  UnifiGem   by first choosing $\mathcal{L}[\cdot]$ as the  Lebesgue norm  and then projecting the corresponding vector fields    onto reproducing kernel Hilbert spaces, please see the supplementary material for a proof.
However, neither SW flow nor MMD flow can model hidden low-dimensional structure with the particle sampling procedure.

\section{Experiments}\label{numerics}
The implementation details on numerical settings, network structures, SGD optimizers, and hyper-parameters are given in the appendix. All experiments are performed using NVIDIA Tesla K80 GPUs. And {The PyTorch code of UnifiGem is available at \url{https://github.com/anonymous/UnifiGem}}.
\subsection{2D Examples}
\begin{figure}[ht!]
\subfigure{
\begin{minipage}[t]{5.2in}
\centering{}
\includegraphics[width=0.7in,height=0.7in]{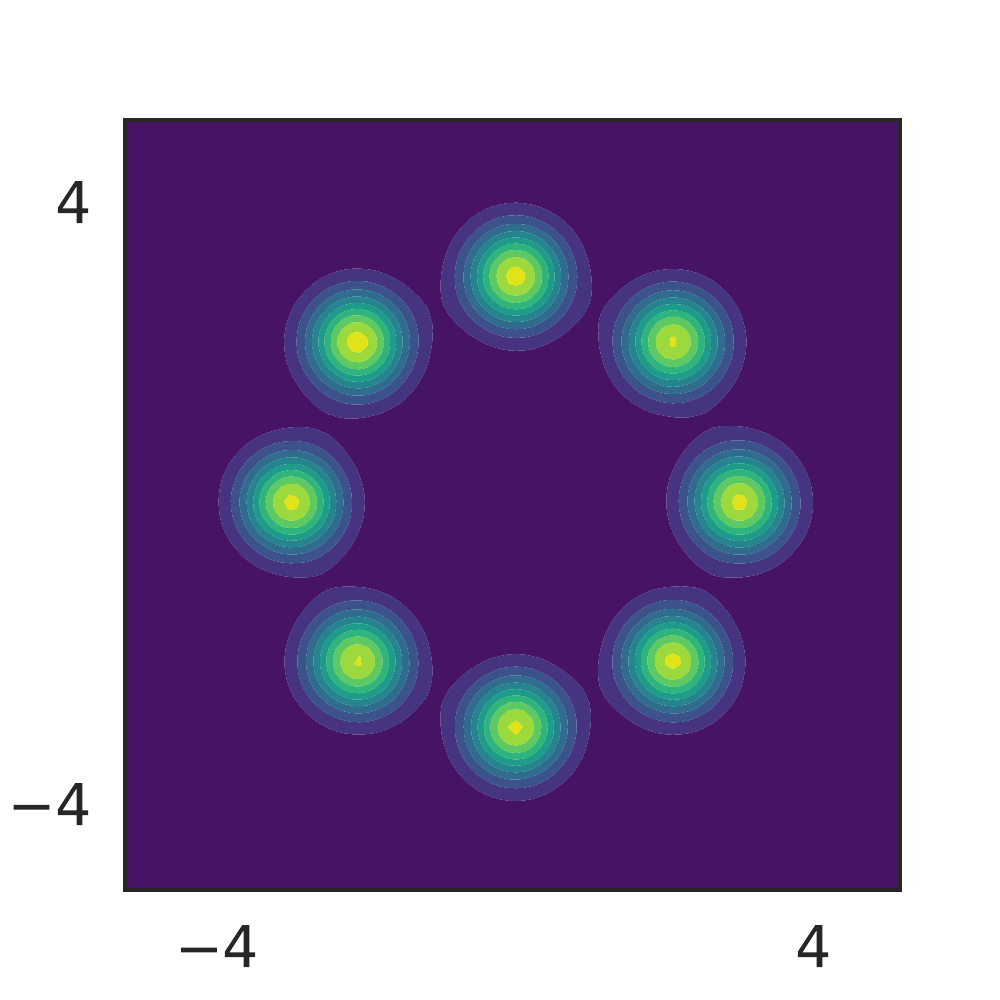}
\includegraphics[width=0.7in,height=0.7in]{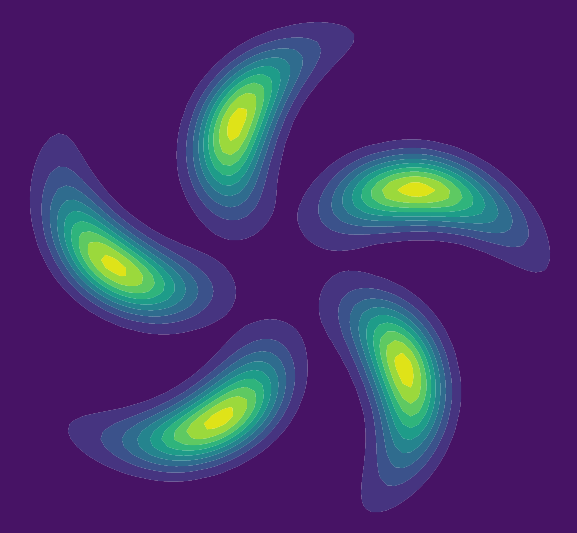}
\includegraphics[width=0.7in,height=0.7in]{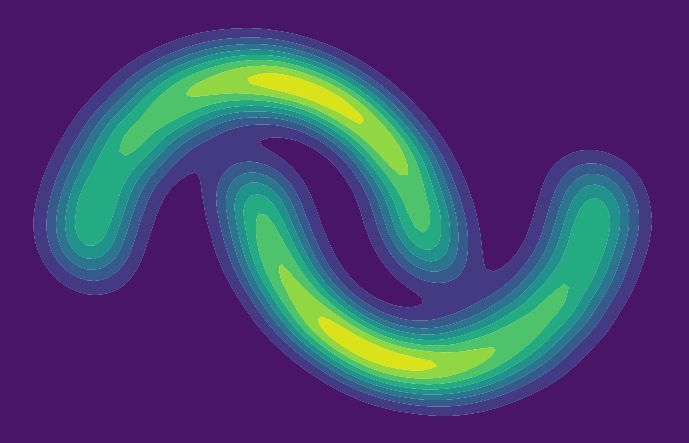}
\includegraphics[width=0.7in,height=0.7in]{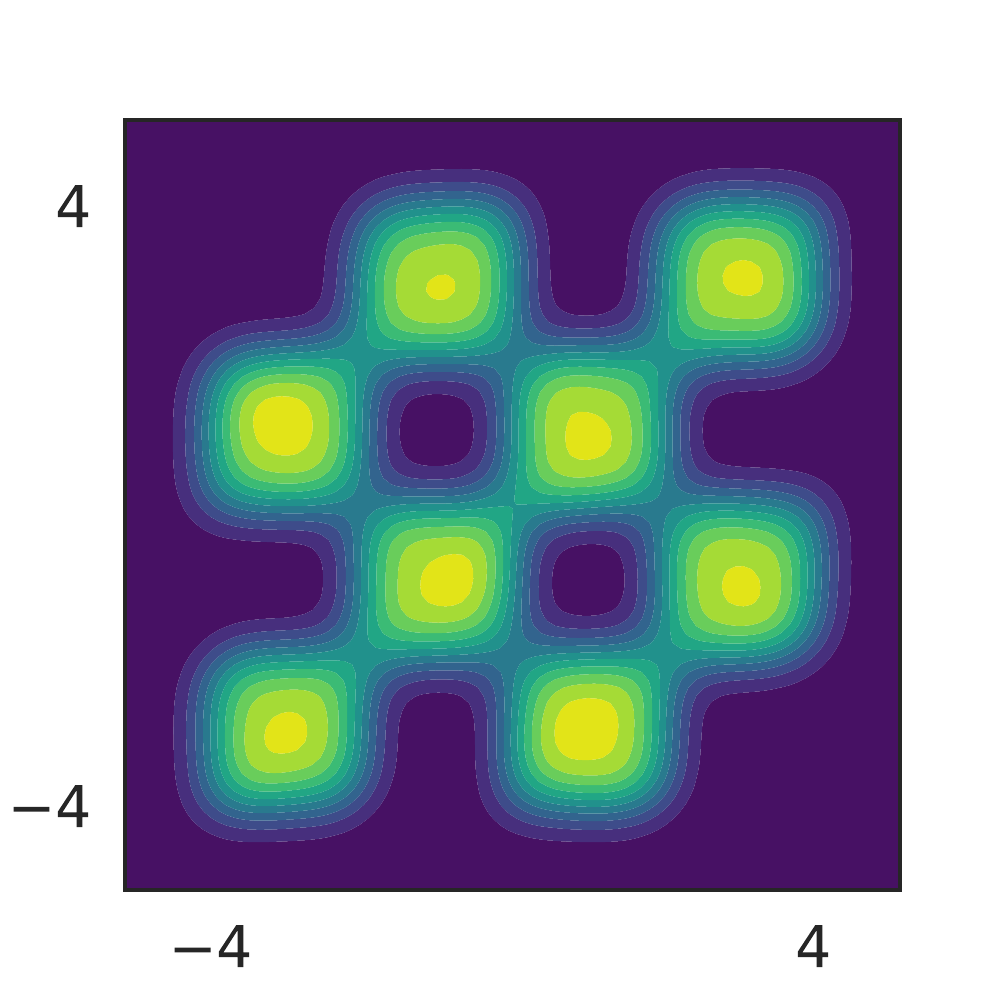}
\includegraphics[width=0.7in,height=0.7in]{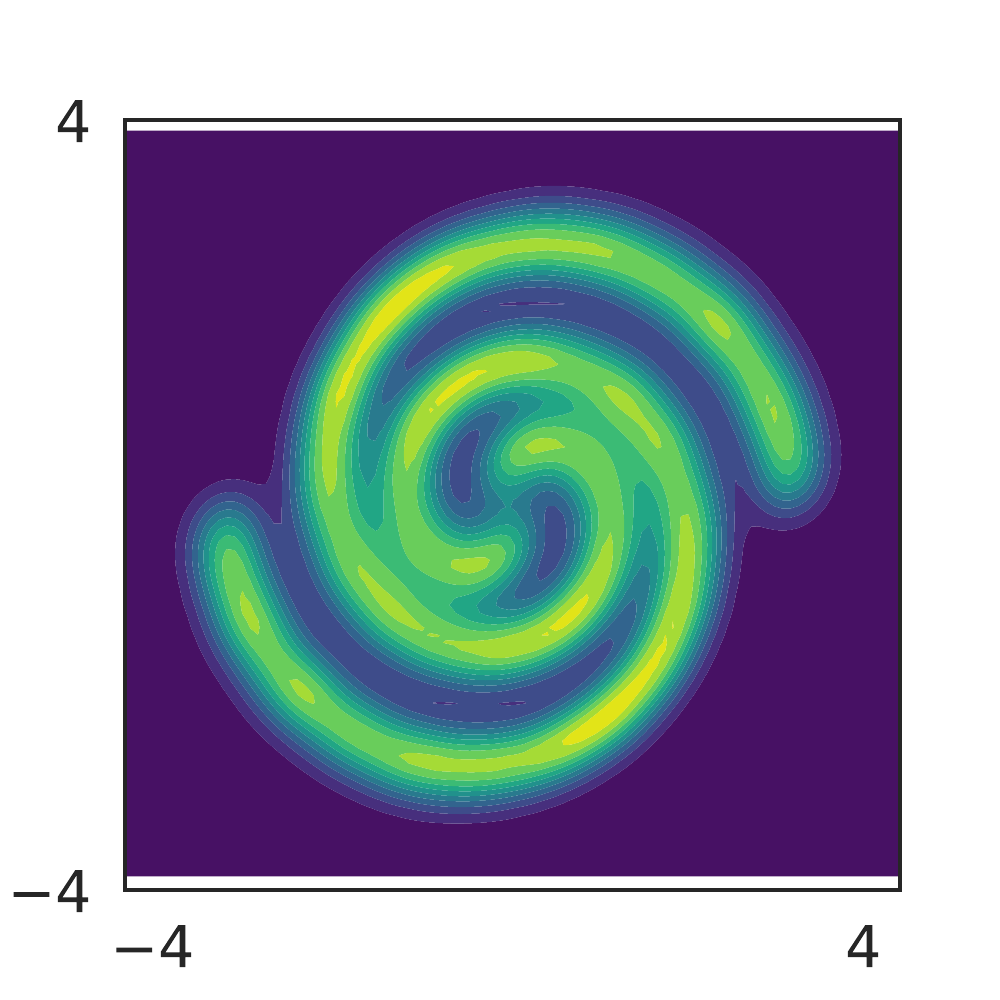}
\includegraphics[width=0.7in,height=0.7in]{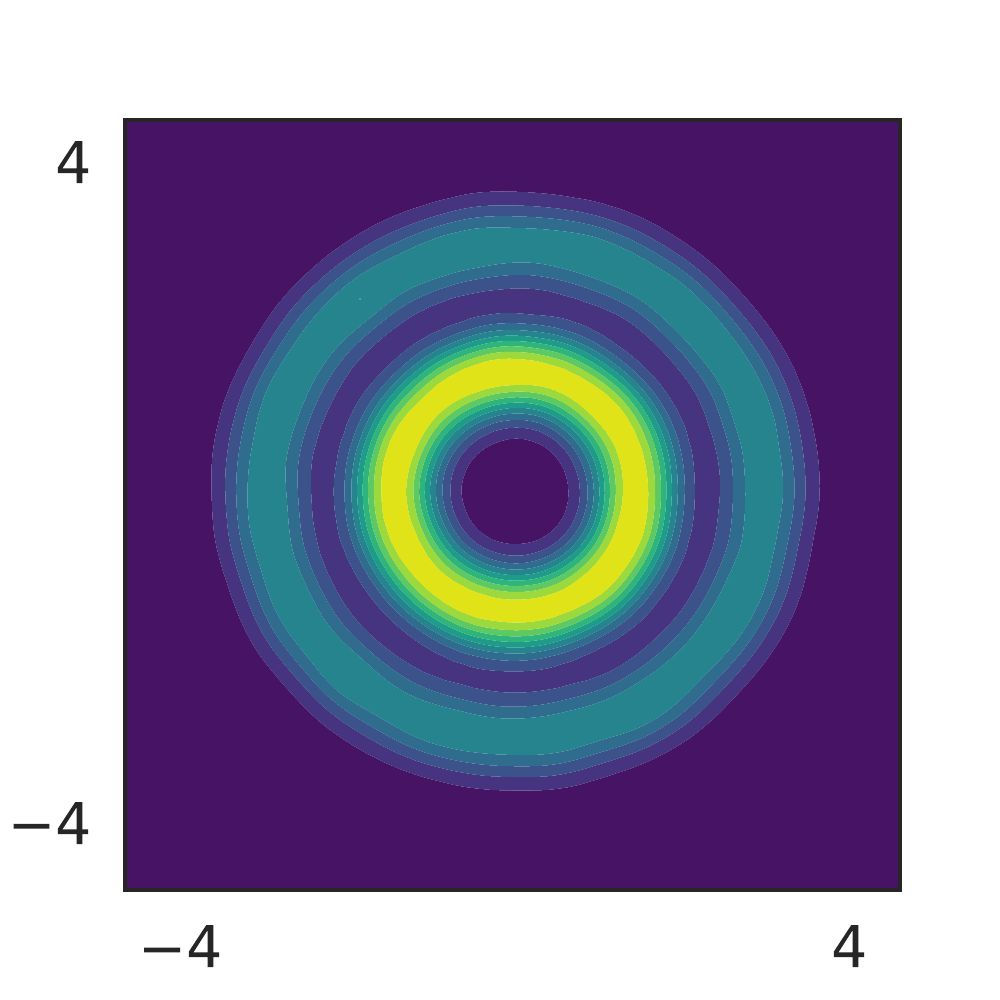}
\end{minipage}
}
\subfigure{
\begin{minipage}[t]{5.2in}
\centering{}
\includegraphics[width=0.7in,height=0.7in]{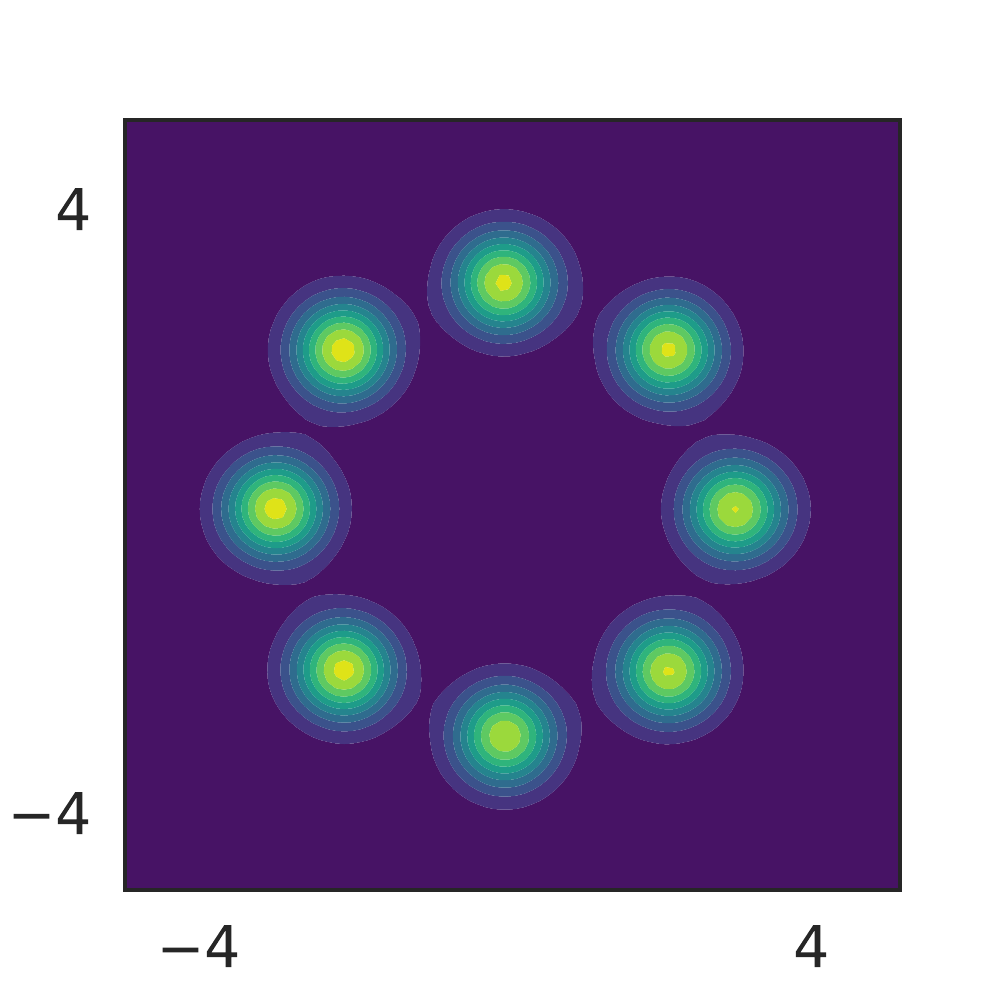}
\includegraphics[width=0.7in,height=0.7in]{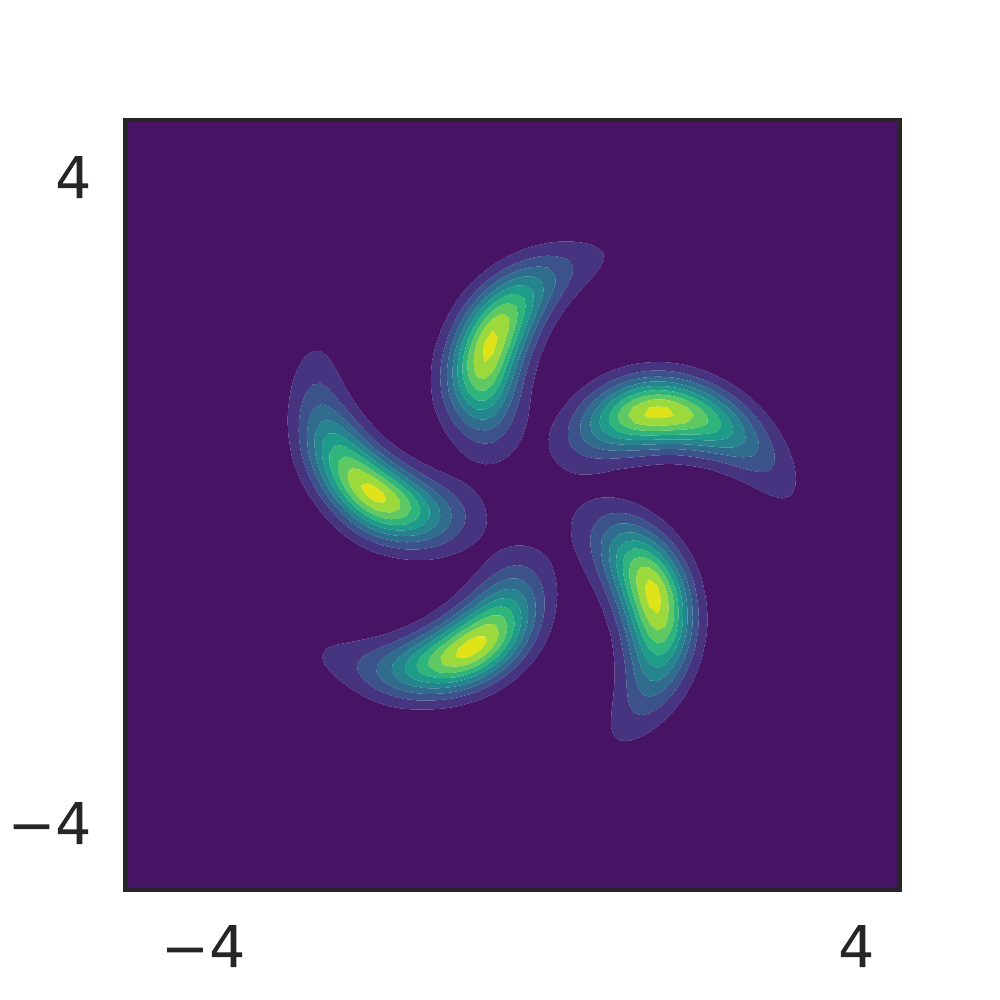}
\includegraphics[width=0.7in,height=0.7in]{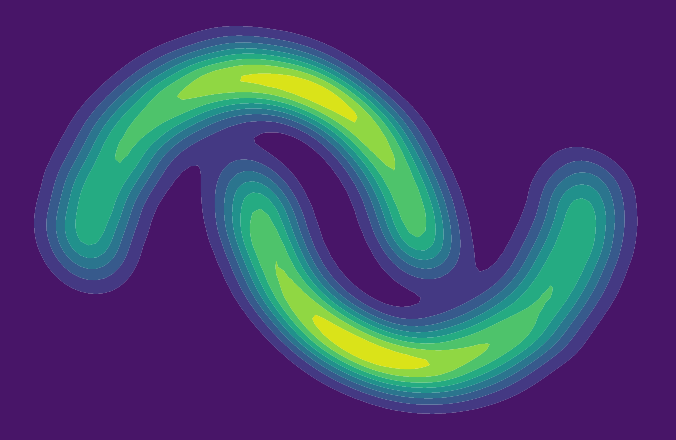}
\includegraphics[width=0.7in,height=0.7in]{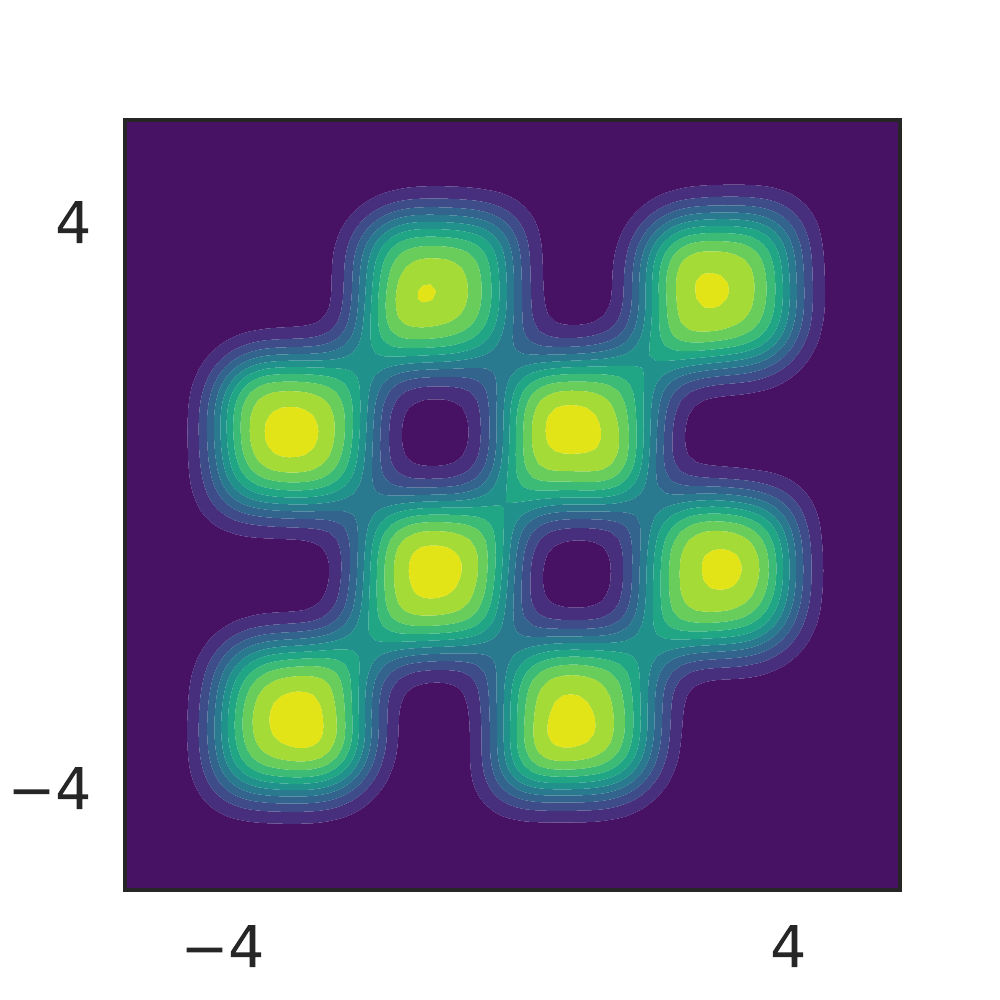}
\includegraphics[width=0.7in,height=0.7in]{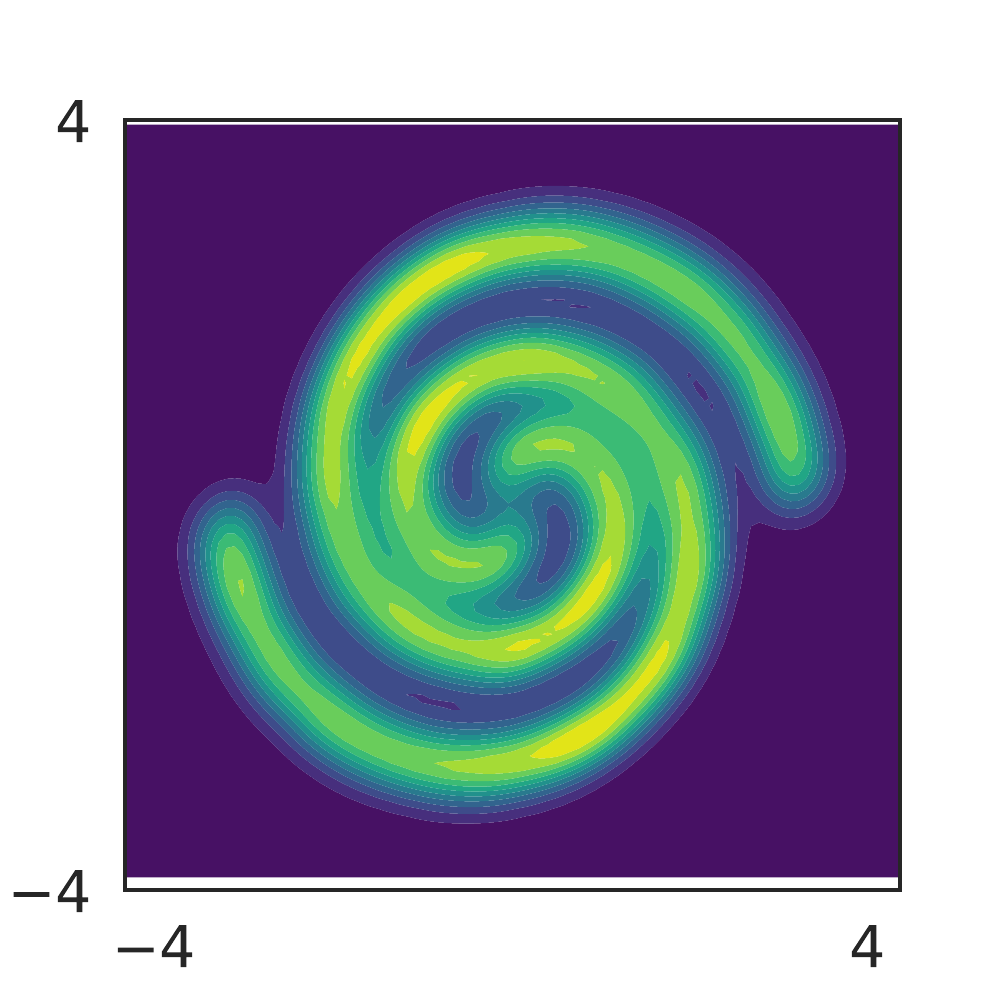}
\includegraphics[width=0.7in,height=0.7in]{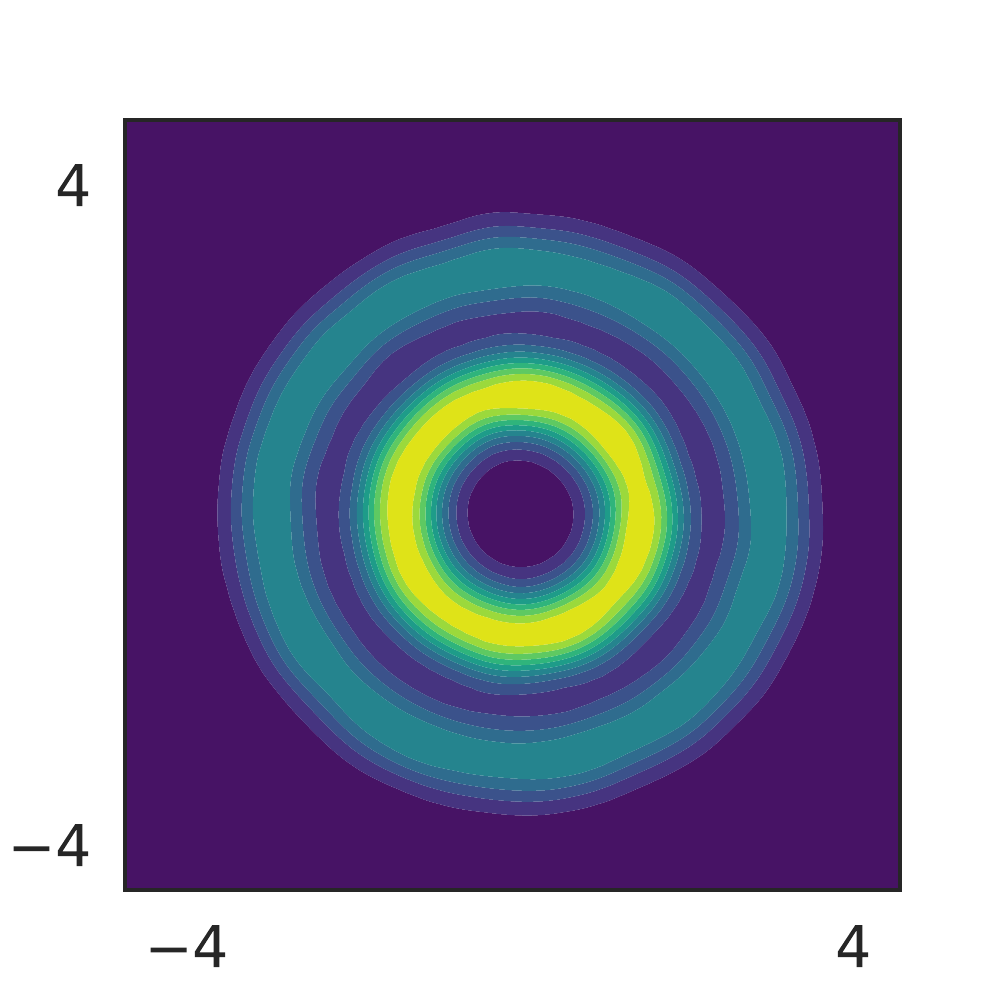}
\end{minipage}
}
\subfigure{
\begin{minipage}[t]{5.2in}
\centering{}
\includegraphics[width=0.7in,height=0.7in]{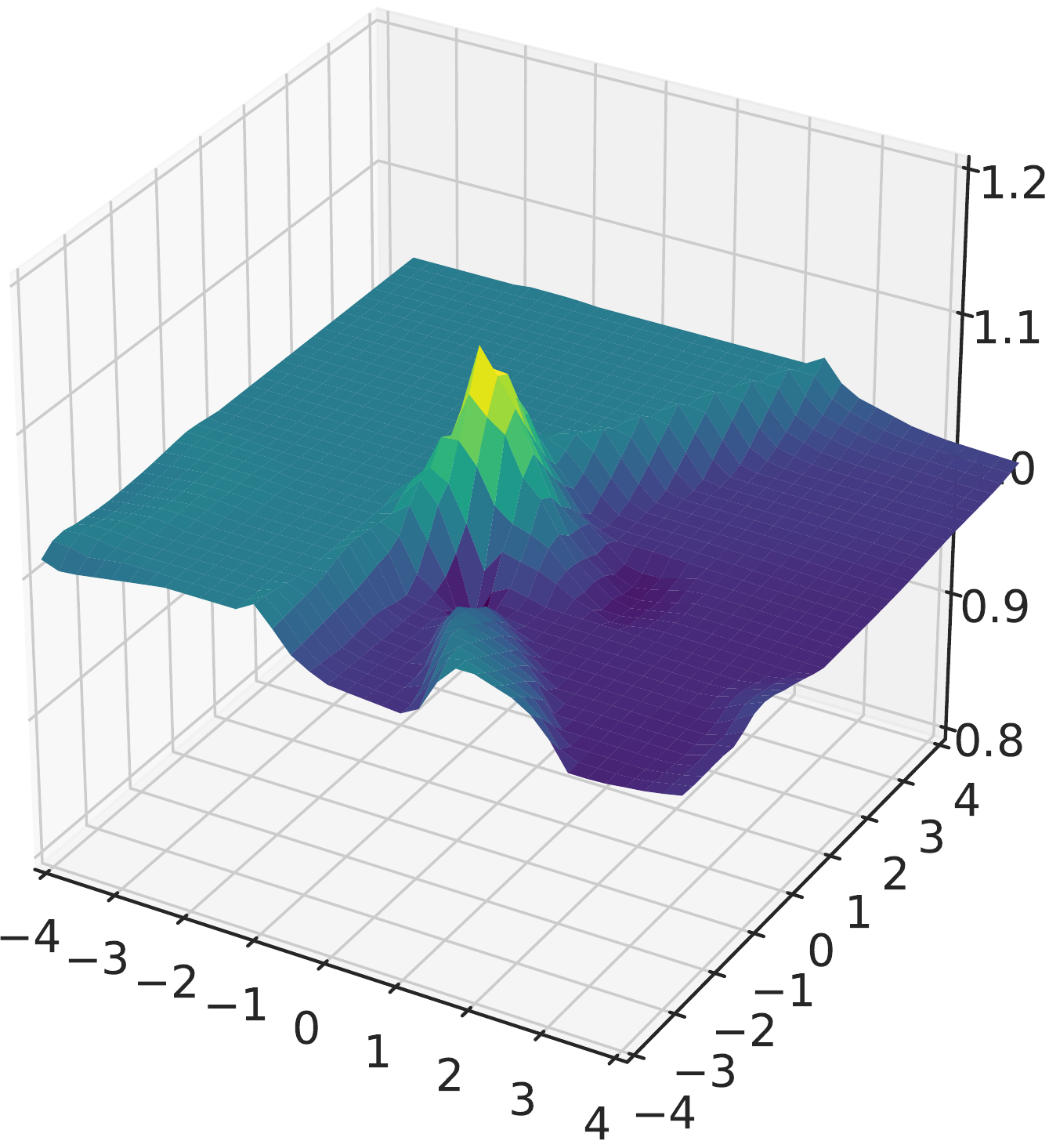}
\includegraphics[width=0.7in,height=0.7in]{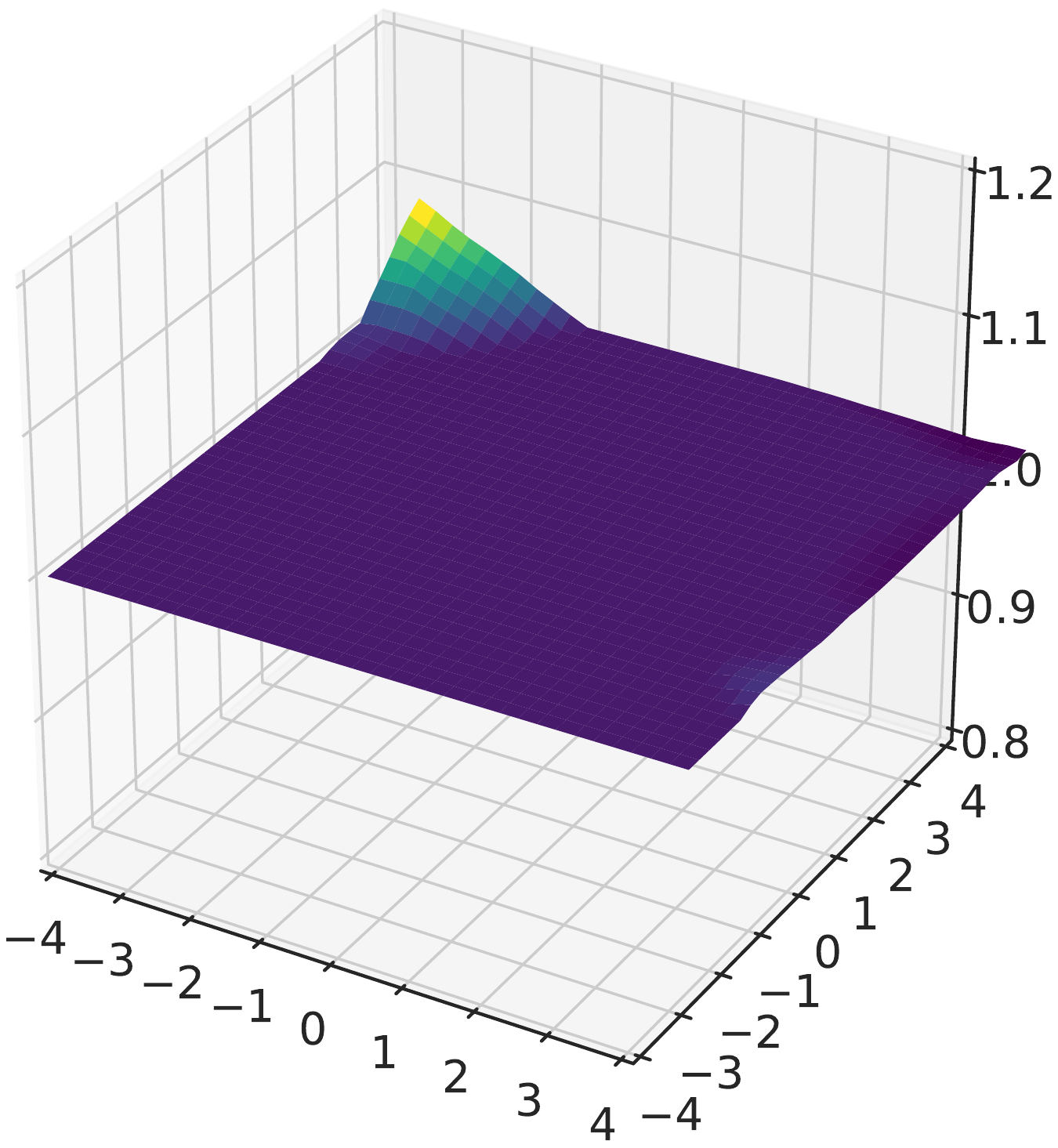}
\includegraphics[width=0.7in,height=0.7in]{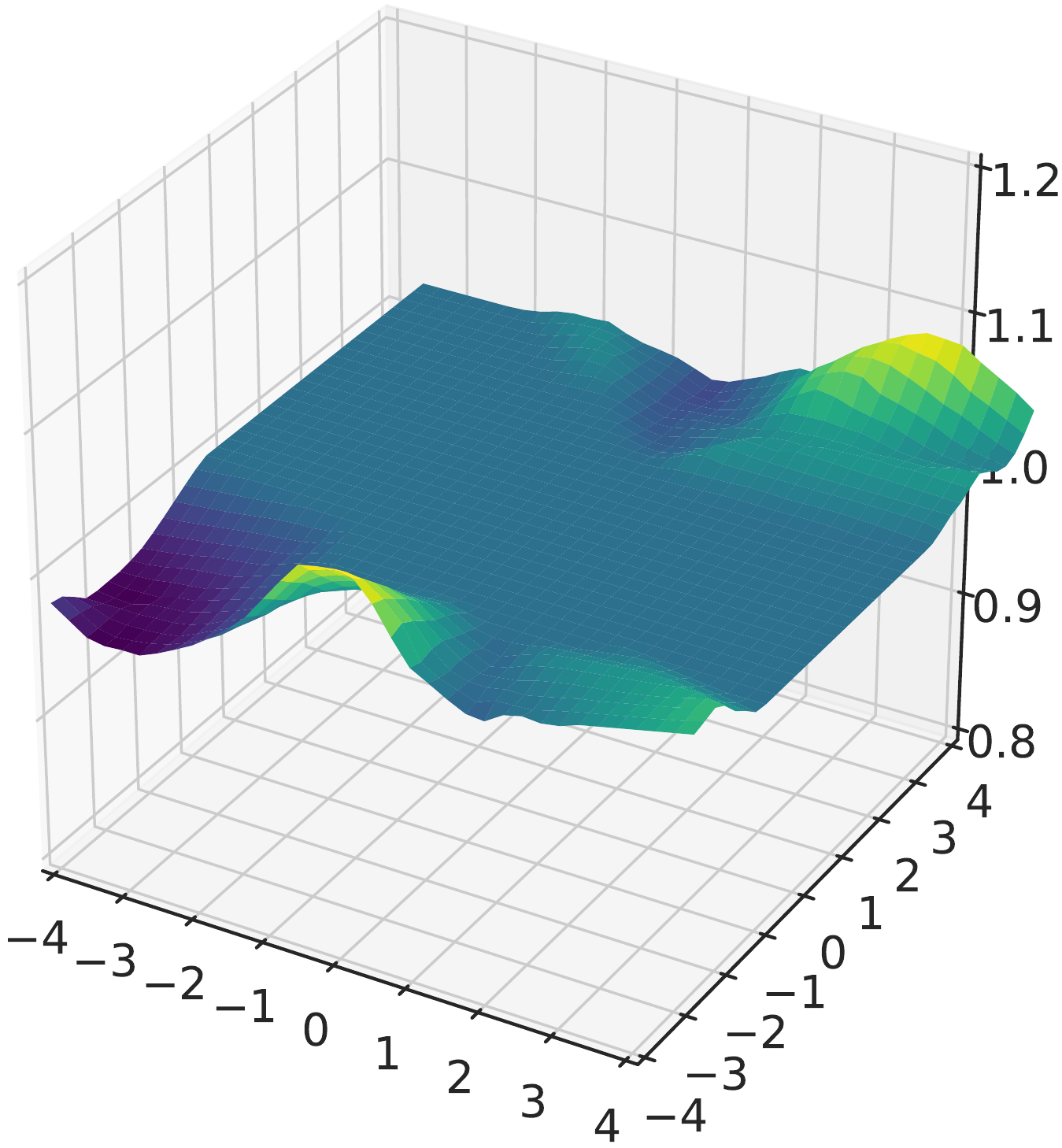}
\includegraphics[width=0.7in,height=0.7in]{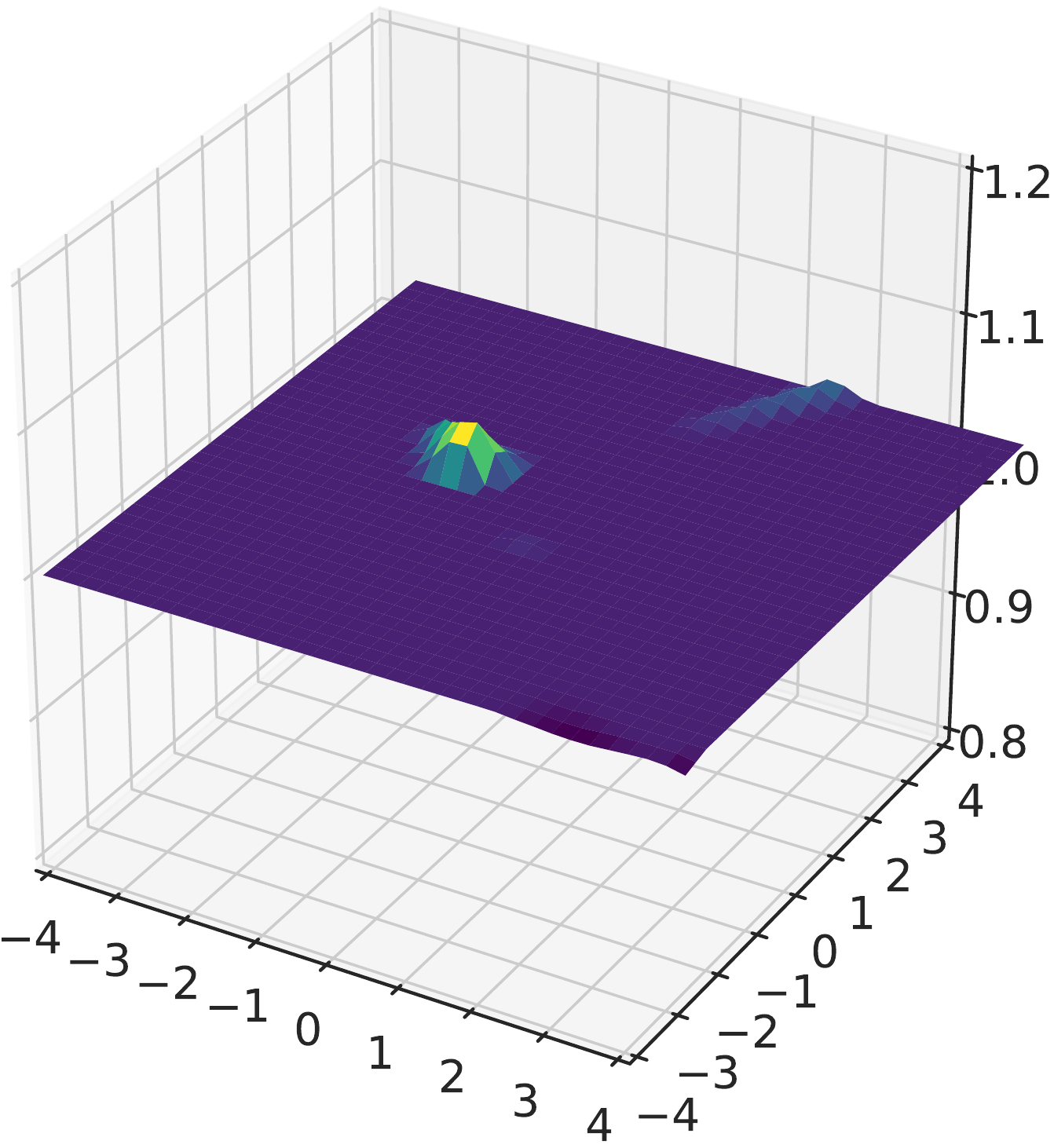}
\includegraphics[width=0.7in,height=0.7in]{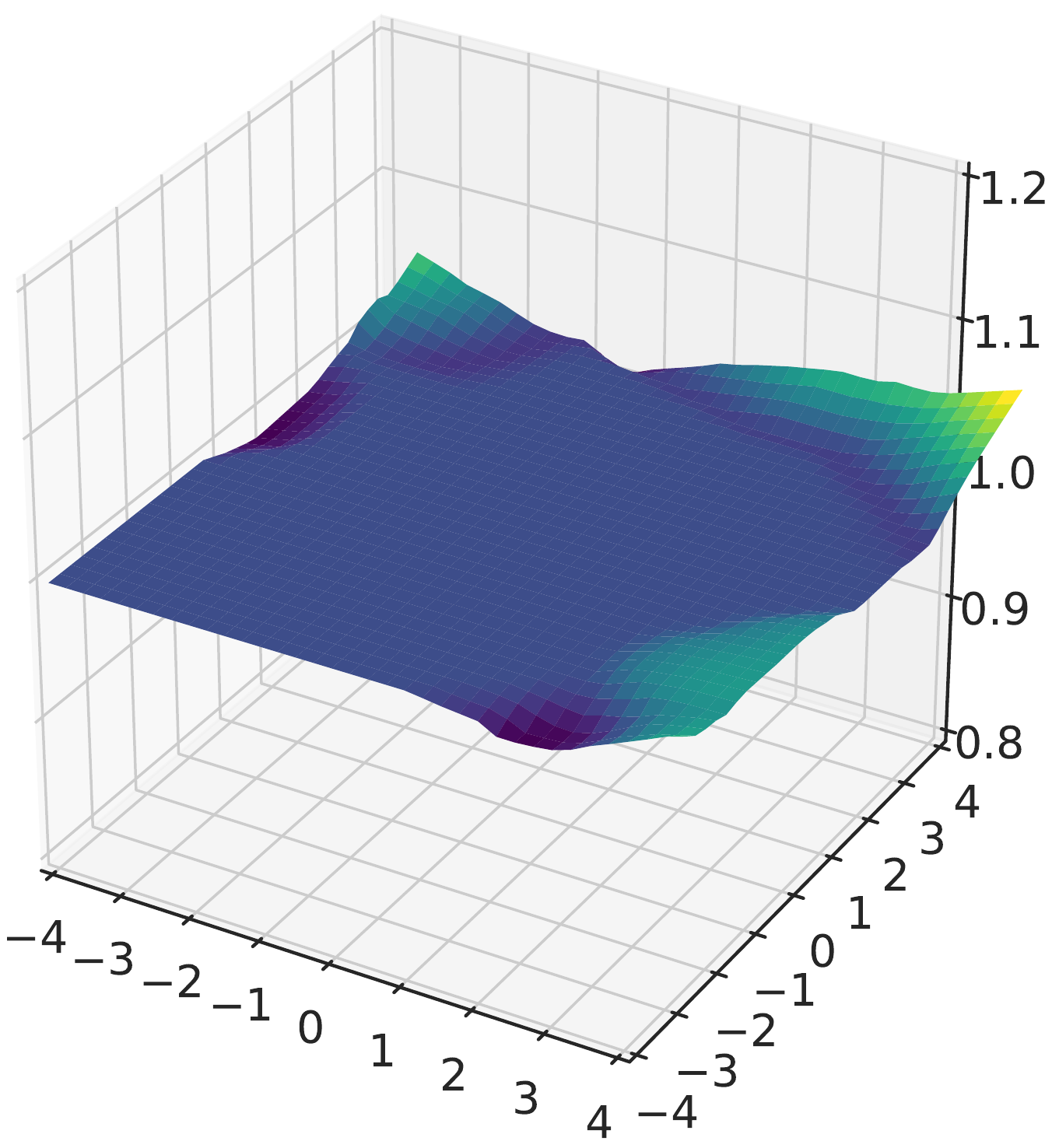}
\includegraphics[width=0.7in,height=0.7in]{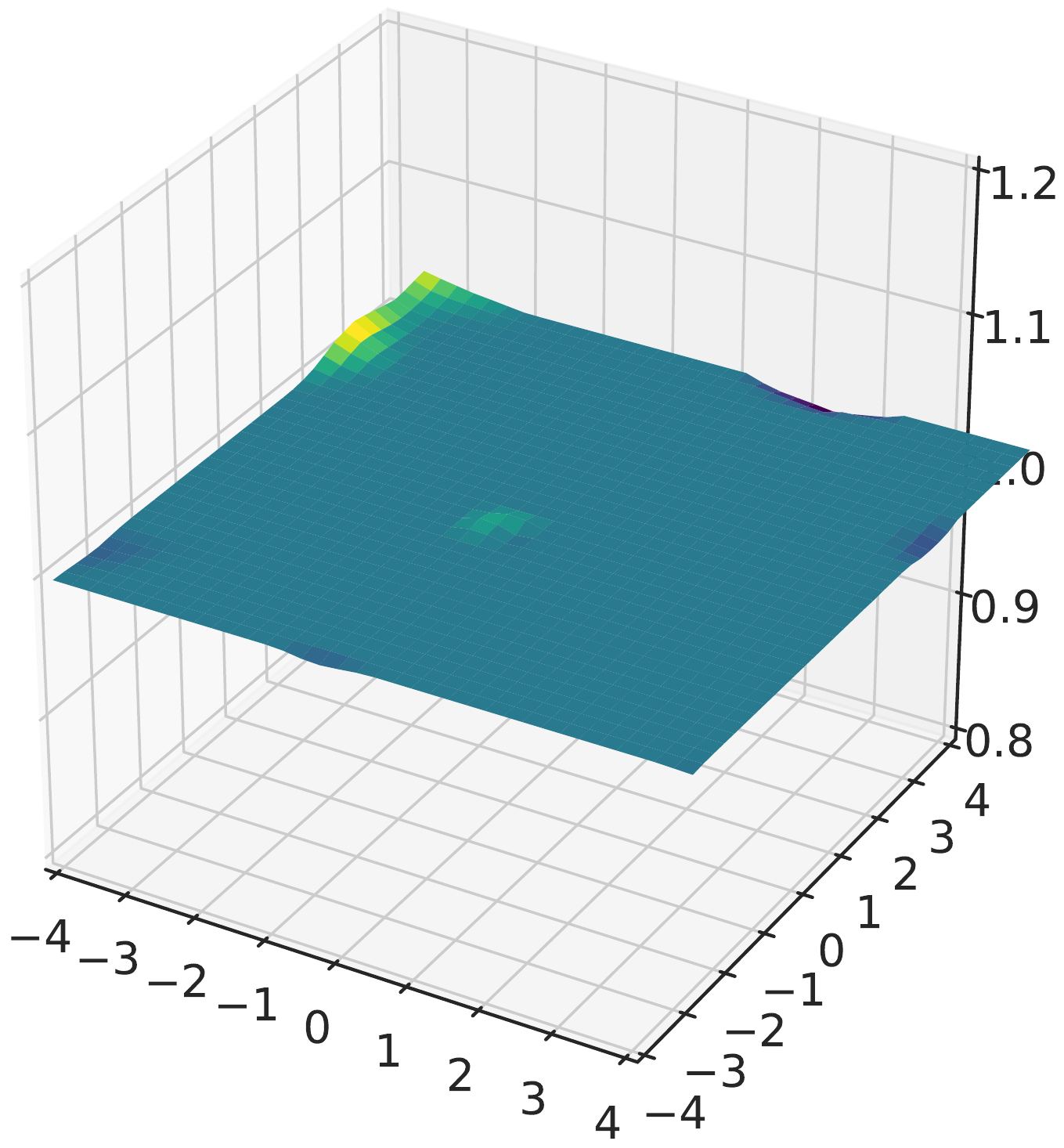}
\end{minipage}
}
\caption{KDE plots of the target samples (the first row) and the corresponding generated samples (the second row). The third row shows surface plots of estimated density ratio after 20k iterations.}
\label{kde}
\end{figure}

We use  UnifiGem to learn 2D  distributions  adapted from \cite{grathwohl2019scalable} with multiple modes and density ridges.
We utilize a multilayer perceptron with ReLU activation
in dynamic deep density-ratio fitting without using gradient penalty.
We use UnifiGem without outer loops to push particles from a predrawn pool consisting of 50k i.i.d. Gaussian particles to evolve in 20k steps.
The first row in Figure \ref{kde} shows kernel density estimation (KDE) plots of 50k samples from target distributions  including (from  left to  right) \emph{8Gaussians, pinwheel, moons, checkerboard, 2spirals,} and \emph{circles}, and the second row shows KDE plots of  the transformed particles via UnifiGem,
and  the third row displays the surface plots of the estimated density-ratio functions  at the end of the  iteration.
As evident by Figure \ref{kde}, KDE plots of generated samples via UnifiGem are nearly indistinguishable from those of the target samples and the estimated density-ratio functions are approximately equal to 1$s$, indicating the learnt distribution  matches the target well.

\begin{figure}[ht!]
\centering{}
\subfigure[\textbf{Left} two figures:  Maps learned without gradient penalty. \textbf{Right} two figures:  Maps learned with  gradient penalty.]{
\begin{minipage}[t]{5.2in}
\centering{}
\includegraphics[width=1.2in,height=1.2in]{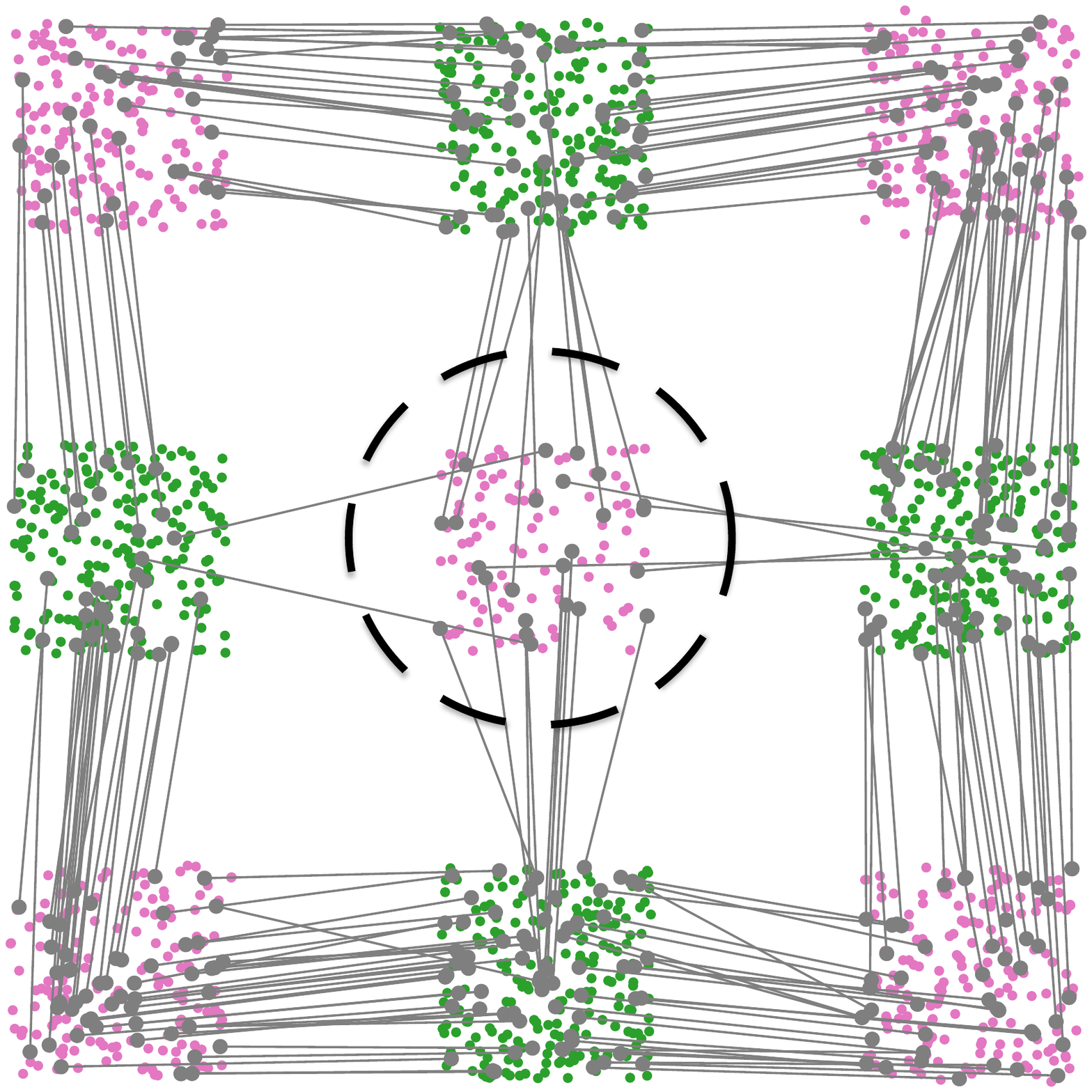}
\includegraphics[width=1.2in,height=1.2in]{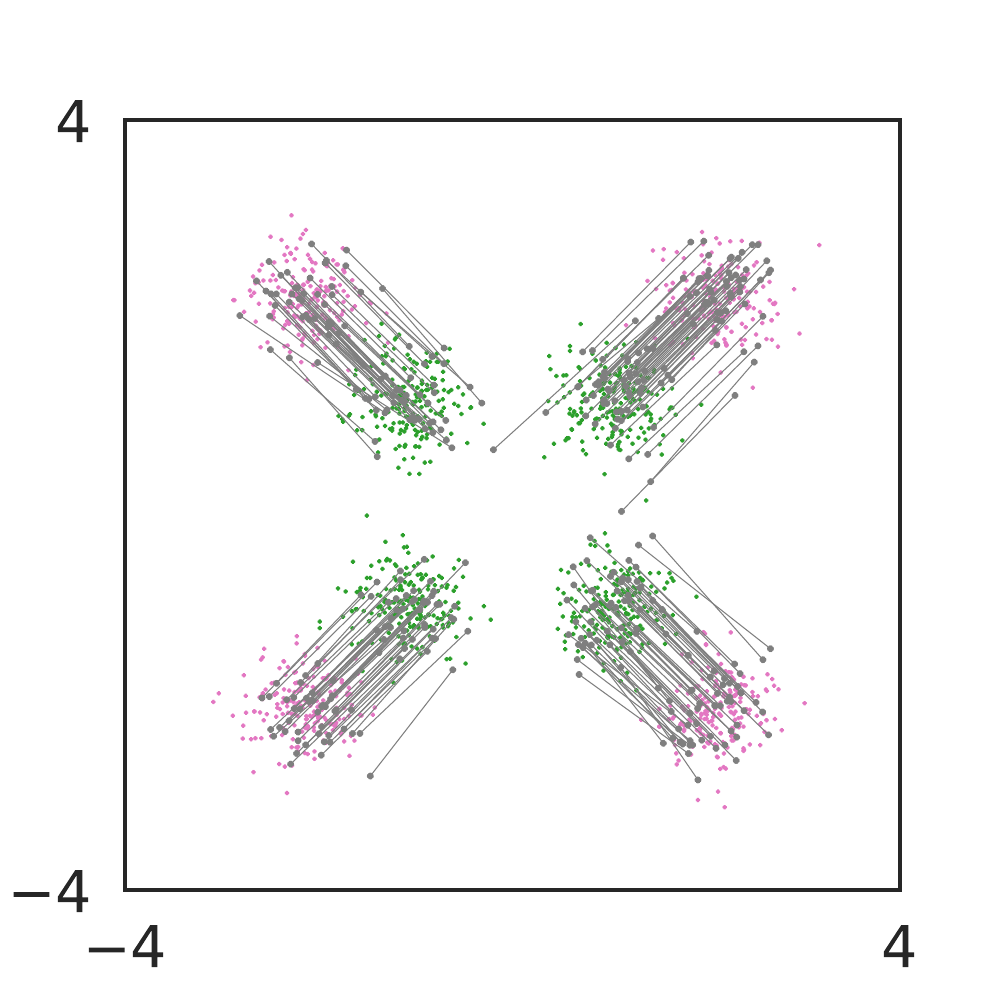}
\includegraphics[width=1.2in,height=1.2in]{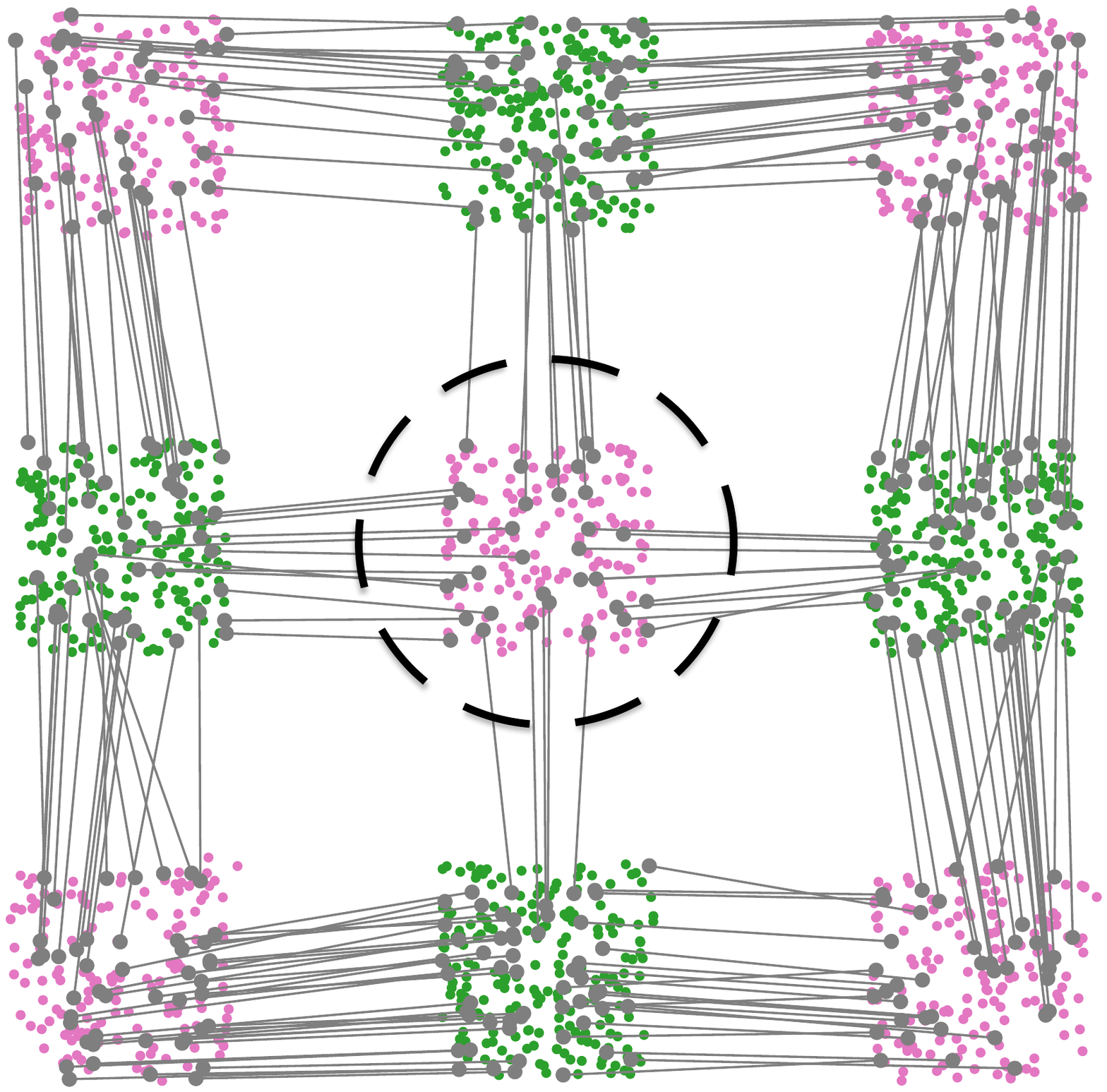}
\includegraphics[width=1.2in,height=1.2in]{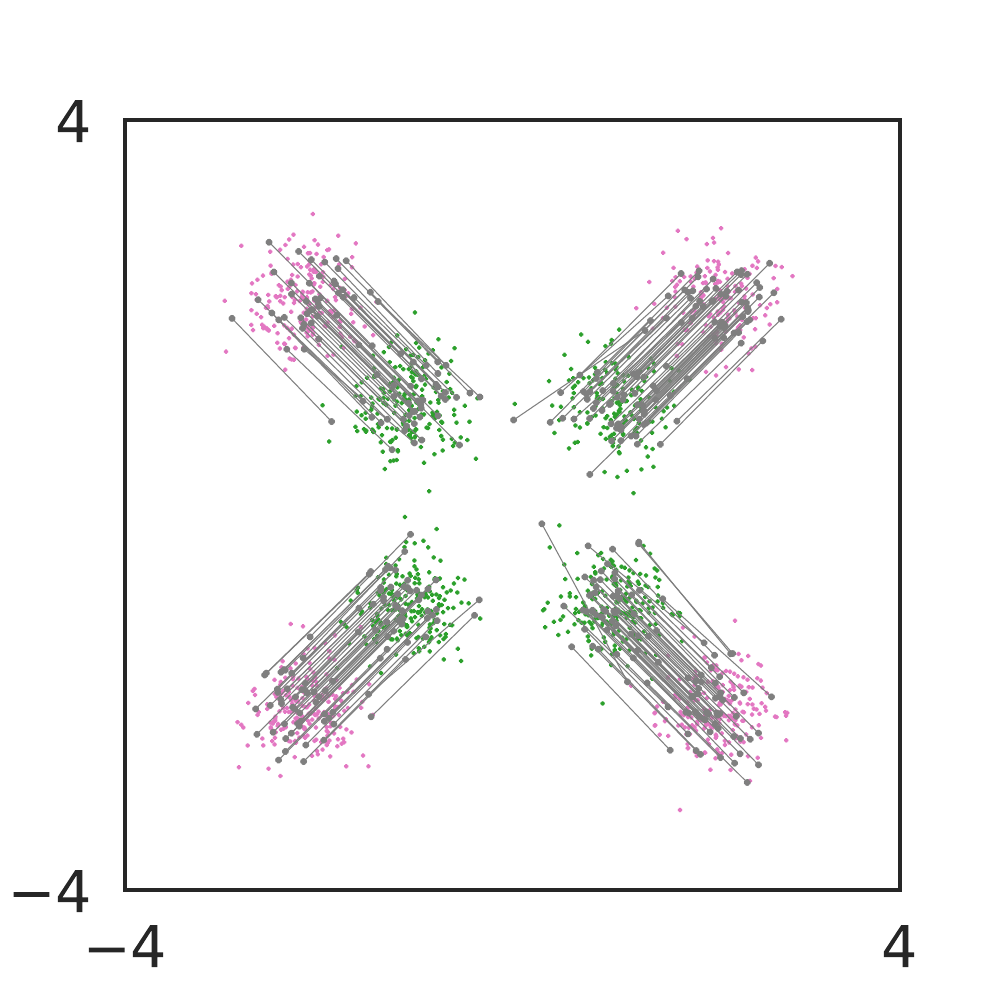}
\end{minipage}
\label{map-map}
}
\subfigure[\textbf{Left} two figures:  Surface plots of estimated density-ratio without gradient penalty. \textbf{Right} two figures:  Surface plots of estimated density-ratio with  gradient penalty.]{
\begin{minipage}[t]{5.2in}
\centering{}
\includegraphics[width=1.2in,height=1.2in]{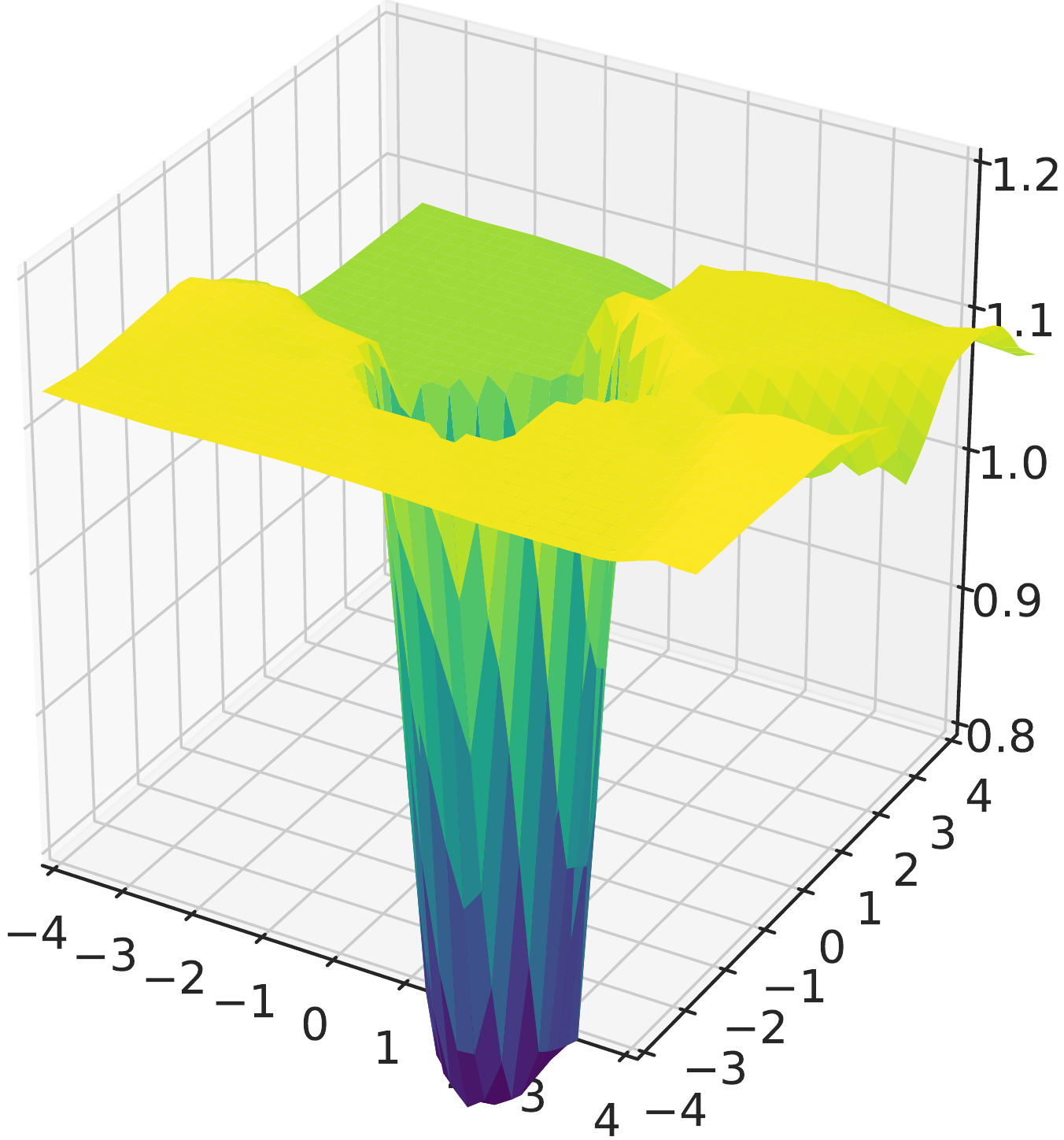}
\includegraphics[width=1.2in,height=1.2in]{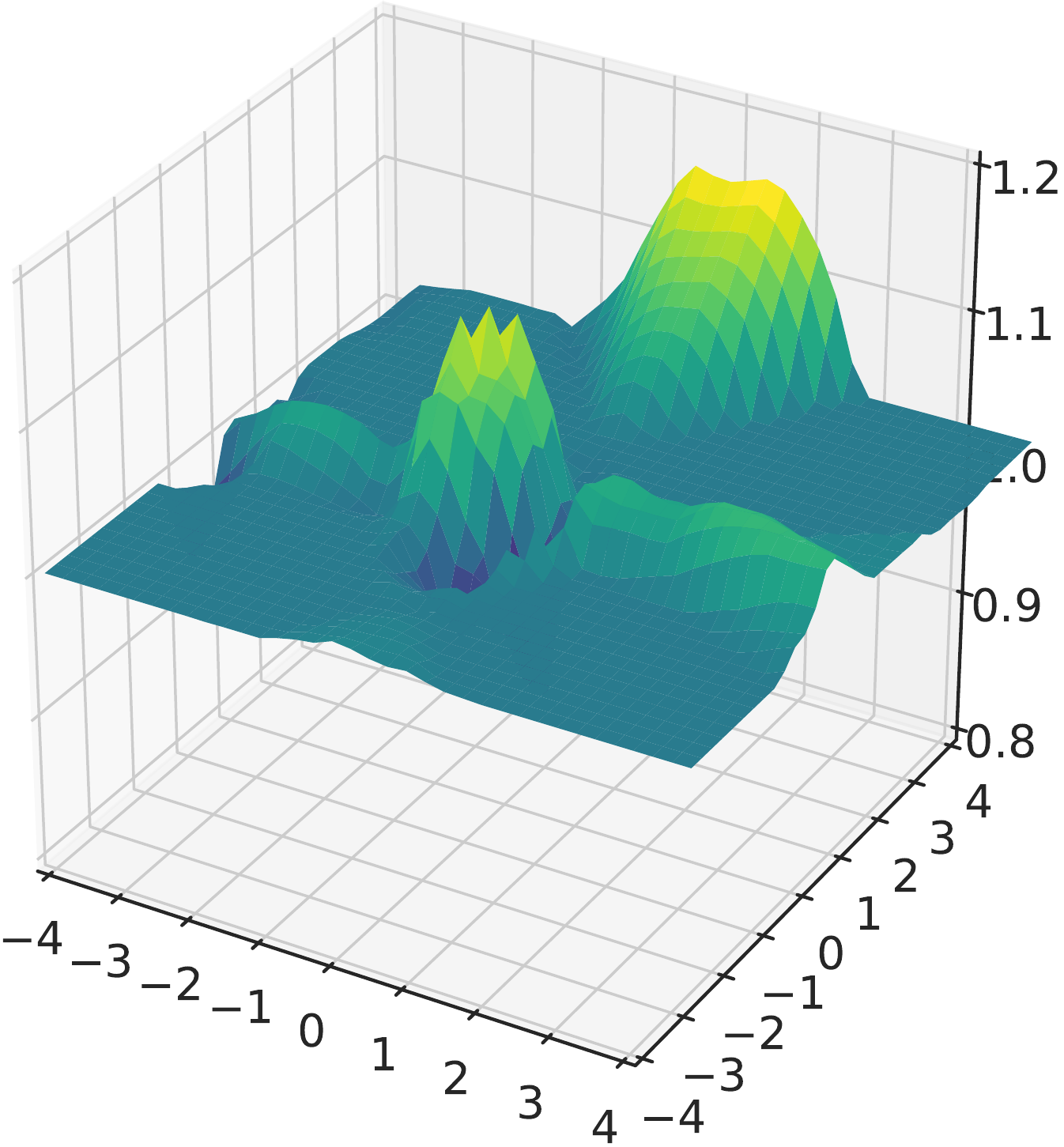}
\includegraphics[width=1.2in,height=1.2in]{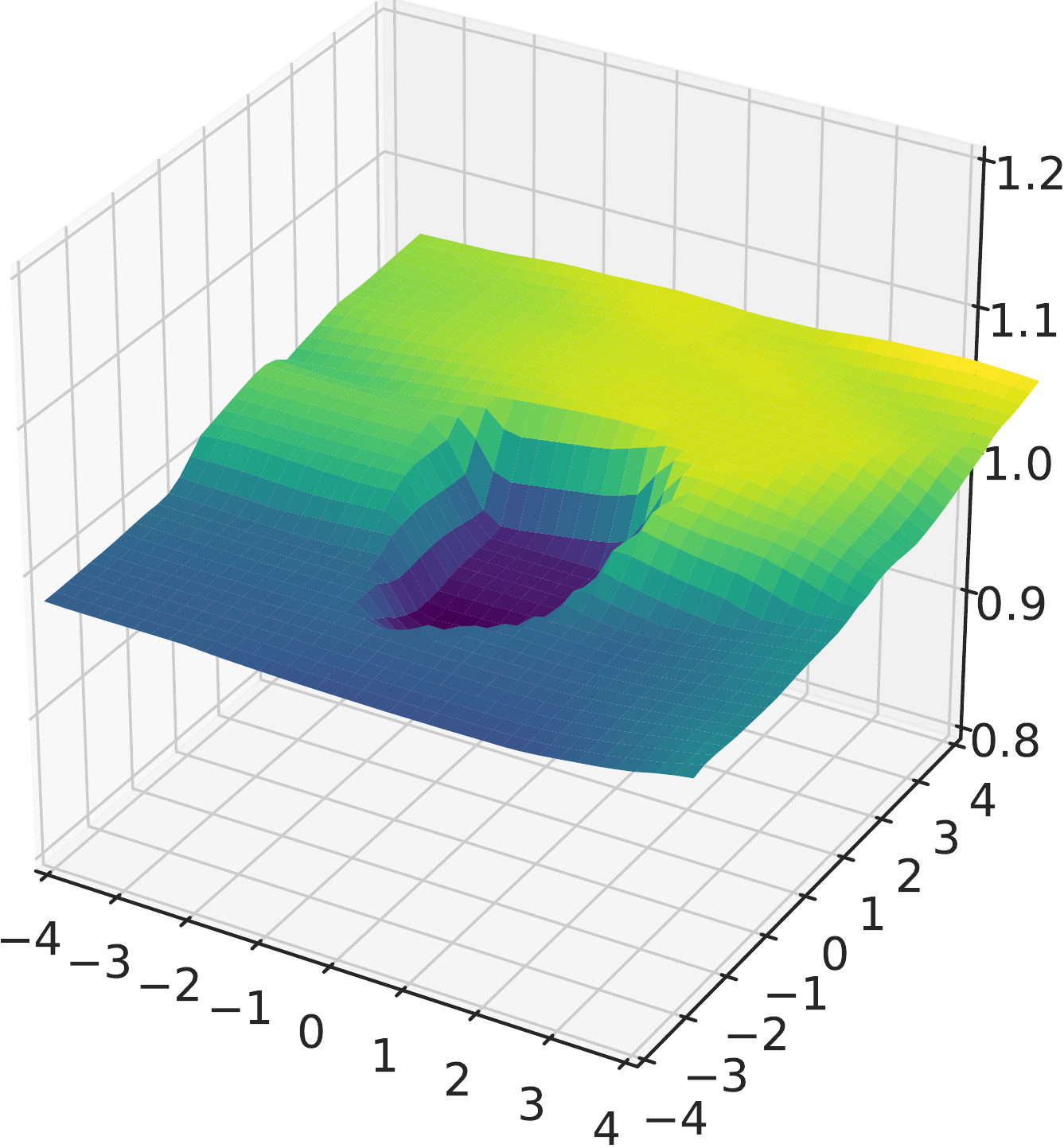}
\includegraphics[width=1.2in,height=1.2in]{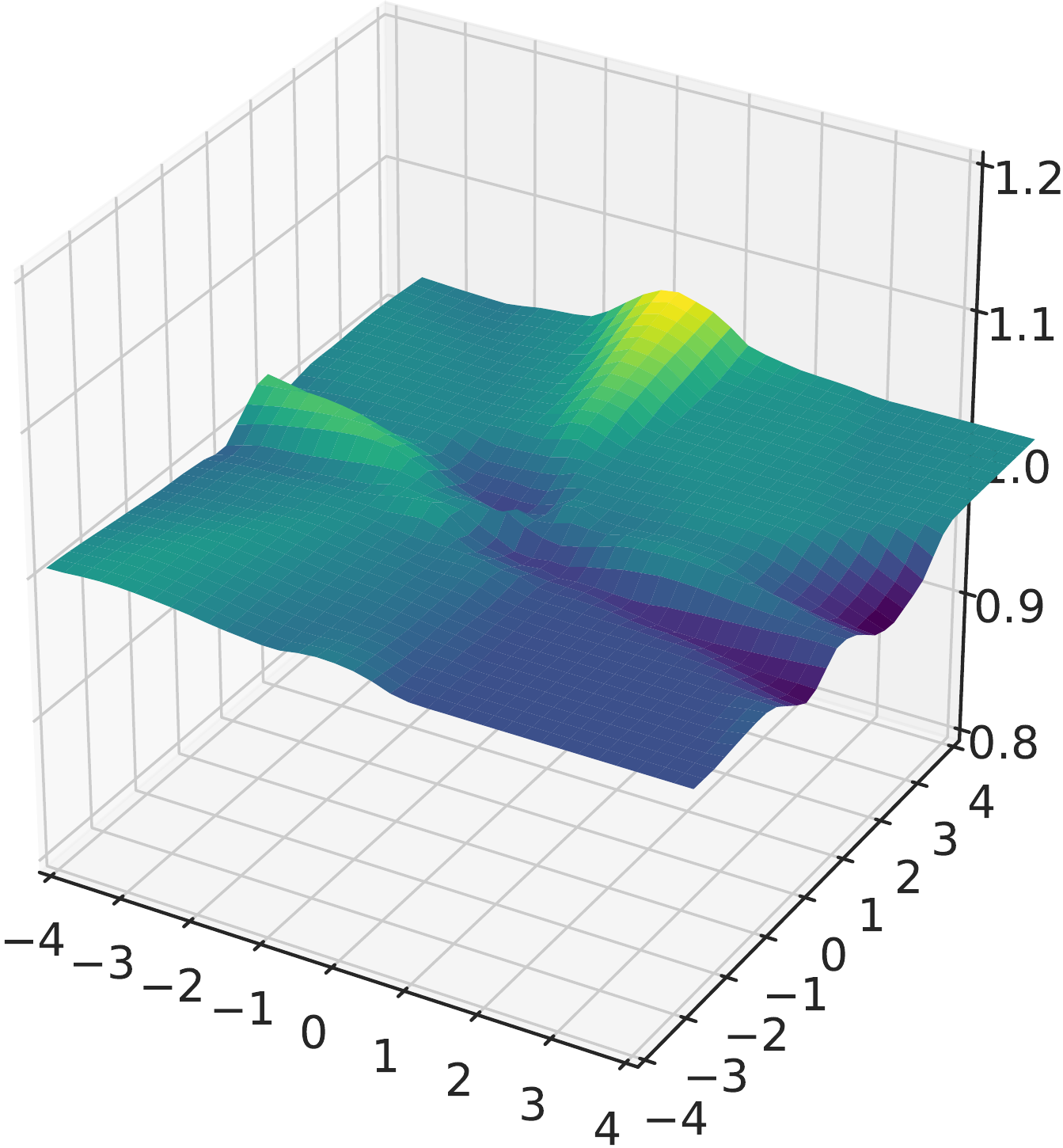}
\end{minipage}
\label{map-dr}
}
\caption{Learned  transport maps and estimated density ratio in  learning  $5squares$ from $4squares$ and learning $large4gaussians$ from $small4gaussians$.}
\label{map}
\end{figure}

Next, we demonstrate the effectiveness of the gradient penalty \eqref{gp} by
visualizing  the transport maps learned in the generative learning  tasks  with the learning targets  $5squares$ and $large4gaussians$ from $4squares$ and $small4gaussians$ respectively.
We use  200 particles connected with grey lines to manifest the learned transport maps.
As shown in  Figure \ref{map-map}, the central squares of $5squares$ were learned better with the gradient penalty, which is consistent with the result of the estimated density-ratio  in Figure \ref{map-dr}.
For $large4gaussians$, the learned transport map exhibited some  optimality under quadratic Wasserstein distance due to the obvious correspondence between the samples in Figure \ref{map-map}, and the   gradient penalty also improves  the density-ratio estimation as expected.

\begin{figure}[ht!]
\includegraphics[width=\linewidth]{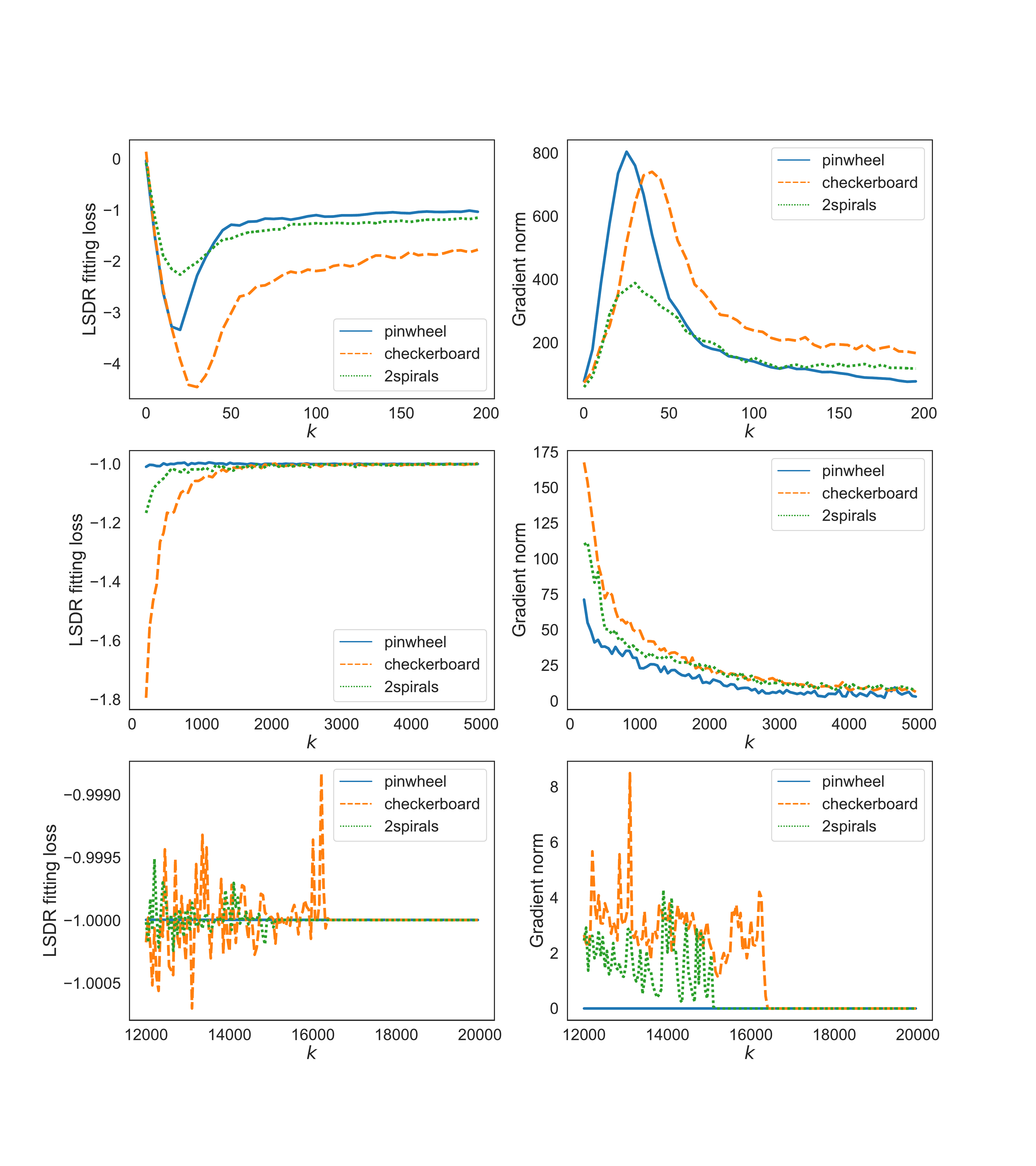}
\centering{}
\caption{Convergence of UnifiGem on \emph{pinwheel, checkerboard} and \emph{2spirals}.
\textbf{Top:} The initialization stage. \textbf{Middle:} The decline stage.
\textbf{Bottom:} The converging stage. \textbf{Left:} LSDR fitting loss \eqref{sf} with $\alpha = 0$.
\textbf{Right:} Estimation of the gradient norm $\mathbb{E}_{X \sim q_k} [\Vert \nabla R_{\phi}(X) \Vert_2]$.}
\label{loss_2d}
\end{figure}

Finally, we illustrate the convergence property of  the learning dynamics of  UnifiGem on synthetic datasets \emph{pinwheel, checkerboard} and \emph{2spirals}.
As shown in Figure \ref{loss_2d}, on the three test datasets, the dynamics of estimated density-ratio fitting losses in \eqref{sf}
share common  patterns for three stages, i.e.,  the initialization stage (top penal), the decline stage (middle panel) and the converging stage (bottom panel).
And both the left panel (LSDR fitting loss \eqref{sf} with $\alpha = 0$)   and the right panel (estimated value of the gradient norm $\mathbb{E}_{X \sim q_k} [\Vert \nabla R_{\phi}(X) \Vert_2]$) demonstrate the estimated LSDR fitting losses in \eqref{sf} (with $\alpha = 0$) converge to the theoretical value $-1$.

\begin{figure}[ht!]
\centering{}
\includegraphics[width=4.0in]{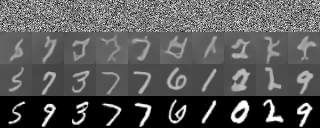} \\
\includegraphics[width=4.0in]{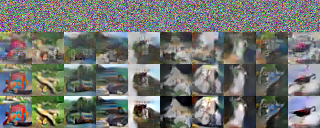}
\caption{Particle evolution of UnifiGem on MNIST and  CIFAR10.}
\label{particle_evol}
\end{figure}

\subsection{Results on benchmark image  data}
\begin{figure}[ht!]
\centering{}
\includegraphics[width=4.0in]{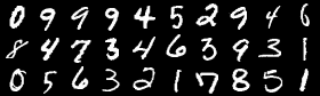} \\
\includegraphics[width=4.0in]{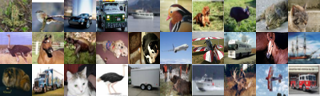} \\
\includegraphics[width=4.0in]{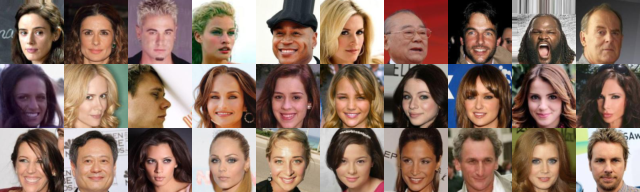} \\
\includegraphics[width=4.0in]{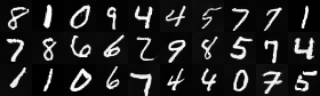} \\
\includegraphics[width=4.0in]{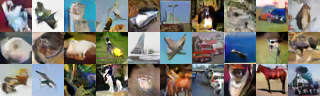}\\
\includegraphics[width=4.0in]{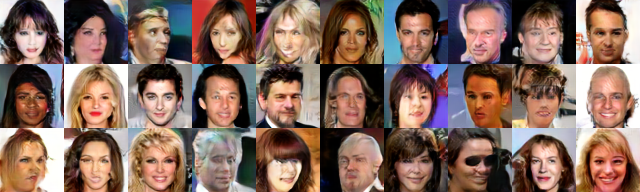}
\caption{Visual comparisons between real images (top 3 panels) and generated images (bottom 3 panels) by  UnifiGem-LSDR-$\chi^2$  on MNIST, CIFAR10 and CelebA.}
\label{sample_comp}
\end{figure}

We show  the performance of applying  UnifiGem  on benchmark image data  MNIST \cite{lecun98},  CIFAR10 \cite{krizhevsky09} and CelebA \cite{liu15}.
The  evolving particles shown in Figure \ref{particle_evol} on  MNIST and CIFAR10 demonstrate that UnifiGem  can  transport samples from a multivariate normal distribution into a target distribution with the same dimension without using the outer loop.
We further compare UnifiGem  using the outer loop  with  state-of-the-art generative models including  WGAN, SNGAN and MMDGAN.
We considered different $f$-divergences, including Pearson $\chi^2$, KL, JS and logD  \cite{gao2019deep} and different deep density-ratio fitting methods (LSDR and LR).
Table \ref{fid-cf10-50k} shows FID  \cite{heusel17} evaluated with  five bootstrap sampling of  UnifiGem with four divergences  on CIFAR10.
We can see that  UnifiGem  attains (usually better)  comparable FID  scores with  the state-of-the-art generative models.
Comparisons of the real samples and learned samples on MNIST,  CIFAR10 and  CelebA are shown in Figure \ref{sample_comp}, where high-fidelity learned samples  are comparable to real samples  visually.

\begin{table}[ht!]
\caption{Mean (standard deviation) of FID scores on  CIFAR10 and results  in last six rows are adapted from \cite{arbel18}.
 }
\label{fid-cf10-50k}
\vskip 0.15in
\begin{center}
\begin{small}
\begin{rm}
\begin{tabular}{lcccr}
\toprule
Models 			& CIFAR10 (50k) \\
\midrule
UnifiGem-LSDR-$\chi^2$ &  \textbf{24.9 (0.1)} \\
UnifiGem-LR-KL 		&  25.9 (0.1) \\
UnifiGem-LR-JS 		&  25.3 (0.1) \\
UnifiGem-LR-logD 		&  \textbf{24.6 (0.1)} \\
\midrule
WGAN-GP			&  31.1 (0.2) \\
MMDGAN-GP-L2		&  31.4 (0.3) \\
SMMDGAN 			&  31.5 (0.4) \\
SN-GAN		  		&  26.7 (0.2) \\
SN-SWGAN 			&  28.5 (0.2) \\
SN-SMMDGAN			&  \textbf{25.0 (0.3)} \\
\bottomrule
\end{tabular}
\end{rm}
\end{small}
\end{center}
\vskip -0.1in
\end{table}

\section{Conclusion and future work} \label{con}
UnifiGem is a unified framework for implicit generative learning via  finding a transport map between a reference distribution and the target distribution.
It is inspired by several fruitful ideas from
optimal transport theory, numerical ODE, density-ratio (density-difference) estimation and deep neural networks. We also provide theoretical guarantees for our proposed approach.
Numerical results on both synthetic datasets and real benchmark datasets support our theoretical findings and demonstrate that  UnifiGem is competitive with the state-of-the-art generative models.

There are two important ingredients in UnifiGem:
the energy functional $\mathcal{L}[\cdot]$ in \eqref{energyFun} and density-ratio (density-difference) estimation.
It can be shown that with a suitable choice of $\mathcal{L}[\cdot]$ and a density-ratio estimation approach, UnifiGem can recover some existing generative models. Thus our theoretical results also provide insights on the properties of these existing methods.
With different combinations  of the energy functionals and density-ratio (density-difference) estimation approach, one can develop new theoretically sound learning procedures under UnifiGem.
It would be interesting to have a through comparison between the procedures resulting from such different combinations. In particular, it is desirable to carefully explore conditions and scenarios of the data structures under which certain choices of the energy functional and density-ratio (density-difference) estimator lead to better performance.

Some aspects and results in this paper are of independent interest. For example, density-ratio estimation is an important problem and of general interest in machine learning and statistics.
The estimation error bound  established in Theorem \ref{th3} for the nonparametric deep density-ratio fitting procedure
is new.
It is a step forward in the direction that shows deep nonparametric estimation can circumvent the curse of dimensionality via exploring the structure of the data \cite{bauer2019deep}.
 It is of interest to use the techniques developed here to study
 deep nonparametric regression and classification.


\section*{Acknowledgements}

The authors are grateful to the anonymous referees, the associate editor and the editor for their helpful comments, which have led to a significant improvement on the quality of the paper.
The work of Jian Huang is supported in part by the NSF grant DMS-1916199.
The work of Y. Jiao was supported in part by
the National Science Foundation of China
under Grant 11871474  and by the research fund
of KLATASDSMOE.
The work of J. Liu is supported by Duke-NUS Graduate Medical School WBS: R913-200-098-263 and MOE2016- T2-2-029 from Ministry of Eduction, Singapore.

\section{Appendix}
In the appendix, we give and
the implementation details on numerical settings, network structures, SGD optimizers, and hyper-parameters in the paper, and  detailed proofs of Lemmas \ref{lem1}-\ref{lem3}, Theorems \ref{th1}-\ref{th3},  Proposition \ref{prop1}-\ref{prop2}, and the proof of  MMD flow being  a special case of  UnifiGem.
\section{Experimental details}
\subsection{2D examples}
Experiments on 2D examples in our work were performed with deep LSDR fitting and the Pearson $\chi^2$ divergence.
For simplicity purposes, outer loops of UnifiGem were omitted and our algorithm became a particle method for approximating solutions of PDEs \cite{chertock2017practical}.
In inner loops, only a multilayer perceptron (MLP) was utilized for dynamic estimation of the density ratio between the model distribution $q_k$ and the target distribution $p$.
The network structure and hyper-parameters in UnifiGem and deep LSDR fitting were shared in all 2D experiments.
We used RMSProp with the learning rate 0.0005 and the batch size 1k as the SGD optimizer.
The details are given in Table \ref{mlp} and Table \ref{param_2D}. We note that $s$ is the step size, $n$ is the number of particles, $\alpha$ is the penalty coefficient, and $T$ is the times of LSDR fitting in each inner loop hereinafter.
\begin{table}[ht!]
\caption{MLP for deep LSDR fitting.}
\label{mlp}
\vskip 0.15in
\begin{center}
\begin{small}
\begin{rm}
\begin{tabular}{lcccr}
\toprule
Layer	& Details 			& Output size \\
\midrule
1		& Linear, ReLU 	& 64 \\
\midrule
2		& Linear, ReLU 	& 64 \\
\midrule
3		& Linear, ReLU 	& 64 \\
\midrule
4		& Linear			& 1 \\
\bottomrule
\end{tabular}
\end{rm}
\end{small}
\end{center}
\vskip -0.1in
\end{table}

\begin{table}[ht!]
\caption{Hyper-parameters in UnifiGem on 2D examples.}
\label{param_2D}
\vskip 0.15in
\begin{center}
\begin{small}
\begin{rm}
\begin{tabular}{lcccr}
\toprule
Parameter		&  $s$ 	& $n$ 	& $\alpha$ 		& $T$ \\
\midrule
Value		& 0.005 	& 50k 	& 0 ${\rm or}$ 0.5 	& 5 \\
\bottomrule
\end{tabular}
\end{rm}
\end{small}
\end{center}
\vskip -0.1in
\end{table}

\subsection{Real image data}

\textbf{Datasets.}
We evaluated UnifiGem on three benchmark datasets including two small datasets MNIST, CIFAR10 and one large dataset CelebA from GAN literature. MNIST contains a training set of 60k examples and a test set of 10k examples as $28\times28$ bilevel images which were resized to $32\times32$ resolution. There are a training set of 50k examples and a test set of 10k examples as $32\times32$ color images in CIFAR10. We randomly divided the 200k celebrity images in CelebA into two sets for training and test according to the ratio 9:1. We also pre-processed CelebA images by first taking a $160\times160$ central crop and then resizing to the $64\times64$ resolution. Only the training sets are used to train our models.

\textbf{Evaluation metrics.}
\emph{Fr\'echet Inception Distance} (FID) \cite{heusel17} computes the Wasserstein distance $\mathcal{W}_2$ with summary statistics (mean $\mu$ and variance $\Sigma$) of real samples $\mathbf{x}s$ and generated samples $\mathbf{g}s$ in the feature space of the Inception-v3 model \cite{szegedy16}, i.e., ${\rm FID} = \Vert \mu_{\mathbf{x}} - \mu_{\mathbf{g}} \Vert^2_2 + {\rm Tr}(\Sigma_{\mathbf{x}} + \Sigma_{\mathbf{g}} - 2(\Sigma_{\mathbf{x}} \Sigma_{\mathbf{g}})^{\frac12})$. Here, FID is reported with the TensorFlow implementation and lower FID is better.

\textbf{Network architectures and hyper-parameter settings.}
We employed the ResNet architectures used by \cite{gao2019deep} in our UnifiGem algorithm.
Especially, the batch normalization \cite{ioffe2015batch} and the spectral normalization \cite{miyato18} of networks were omitted for UnifiGem-LSDR-$\chi^2$.
To train neural networks, we set SGD optimizers as RMSProp with the learning rate 0.0001 and the batch size 100.
Inputs $\{ Z_i \}_{i=1}^n$ in UnifiGem with outer loops were vectors generated from a 128-dimensional standard normal distribution on all three datasets. Hyper-parameters are listed in Table \ref{param_ol} where $IL$ expresses the number of inner loops in each outer loop. Even without outer loops, UnifiGem can generate images on MNIST and CIFAR10 as well by making use of a large set of particles. Table \ref{param_no_ol} shows the hyper-parameters.

\begin{table}[ht!]
\caption{Hyper-parameters in UnifiGem \textbf{with} outer loops on real image datasets.}
\label{param_ol}
\vskip 0.15in
\begin{center}
\begin{small}
\begin{rm}
\begin{tabular}{lcccccr}
\toprule
Parameter		& $\ell$ 	& $s$ 	& $n$ 	& $\alpha$ 	& $T$ 	& $IL$ \\
\midrule
Value 		& 128  	& 0.5  	& 2k 		& 0 			& 1 		& 20 \\
\bottomrule
\end{tabular}
\end{rm}
\end{small}
\end{center}
\vskip -0.1in
\end{table}

\begin{table}[ht!]
\caption{Hyper-parameters in UnifiGem \textbf{without} outer loops on real image datasets.}
\label{param_no_ol}
\vskip 0.15in
\begin{center}
\begin{small}
\begin{rm}
\begin{tabular}{lcccr}
\toprule
Parameter		&  $s$  	& $n$ 	& $\alpha$ 	& $T$ \\
\midrule
Value		& 0.5  	& 4k 		& 0 			& 5 \\
\bottomrule
\end{tabular}
\end{rm}
\end{small}
\end{center}
\vskip -0.1in
\end{table}

\subsection{Proof of Lemma \ref{lem1}}
\begin{proof}
This is well known results \cite{brenier1991polar,mccann1995existence,santambrogio2015optimal}, see, Section 1.7.6 on page 54 of \cite{santambrogio2015optimal} for example.
\end{proof}
\subsection{Proof of Theorem \ref{th1}}
\begin{proof}
We show the results item by item.
(i) The continuity equation (12) follows from the definition  of the gradient flow directly, see, page 281 in \cite{ambrosio2008gradient} for detail. \\
(ii)  Recall  $\mathcal{L}[\mu]$ is a functional  on  $\mathcal{P}_2^{a}(\mathbb{R}^{m})$. By the classical results in calculus  of variation \cite{gelfand2000calculus}, $$\frac{\partial \mathcal{L}[q]}{\partial q}(\x) = \frac{\mathrm{d}}{\mathrm{d}t} \mathcal{L}[q+tg]\mid_{t = 0} = F^{\prime}(q(\x)),$$  where, $\frac{\partial \mathcal{L}[q]}{\partial q} $ denotes the first order of variation of $\mathcal{L}[\cdot]$ at $q$, and $q, g $ are the densities  of $\mu$ and an arbitrary  $ \xi \in \mathcal{P}_2^{a}(\mathbb{R}^{m})$, respectively. Let
$$L_{F}(z) = z F^{\prime}(z) -F(z): \mathbb{R}^{1} \rightarrow \mathbb{R}^{1}.$$
Some algebra shows, $$\nabla L_{F}(q(\x)) = q(\x) \nabla F^{\prime} (q(\x)).$$ Then, it follows from Theorem 10.4.6 in \cite{ambrosio2008gradient} that $$ \nabla F^{\prime} (q(\x)) = \partial^{o}{L}(\mu),$$ where, $\partial^{o}{L}(\mu)$ denotes the one in $\partial{L}(\mu)$ with minimum length.  The above display  and  the definition of  gradient flow  implies  the representation of the velocity fields $\mathbf{v}_t$.\\
(iii) The first equality  follows from chain rule and integration by part, see, Theorem 24.2 in \cite{villani2008optimal} for detail.
The second one on linear convergence  follows from Theorem 24.7 in \cite{villani2008optimal},  where the assumption on $\lambda$  in equation (24.6) is equivalent to the $\lambda$-geodetically convex assumption here.\\
(iv) Similar to (i) see, page 281 in \cite{ambrosio2008gradient} for detail.
\end{proof}
\subsection{Proof of Proposition \ref{prop1}}
\begin{proof}
The time dependent form of (7)-(8) reads
\begin{align*}
\frac{\mathrm{d} \x_t}{\mathrm{d} t} &= \nabla \Phi_t(\x_t), \ \  \mathrm{with} \ \  \x_0  \sim q,\\
\frac{\mathrm{d} \ln  q_t(\x_t)}{\mathrm{d} t} &= - \Delta
\Phi_t (\x_t), \ \  \mathrm{with} \ \  q_0 = q.
\end{align*}
By chain rule and substituting the first equation into the second one, we have
 \begin{align*}
 \frac{1}{q_t}(\frac{\mathrm{d}q_t}{\mathrm{d}t}+\frac{\mathrm{d}q_t}{\mathrm{d}\x_t}\frac{\mathrm{d}\x_t}{\mathrm{d}t}) &= \frac{1}{q_t}(\frac{\mathrm{d}q_t}{\mathrm{d}t}+\nabla q_t\nabla \Phi_t(\x_t))\\
 &=- \Delta\Phi_t (\x_t),
\end{align*}
which implies,
$$\frac{\mathrm{d}q_t}{\mathrm{d}t} = - q_t\Delta\Phi_t (\x_t) -\nabla q_t\nabla \Phi_t(\x_t) = -\nabla\cdot(q_t \nabla \Phi_t).$$
By (13), the above display coincides with the continuity equation  (12) with $\mathbf{v}_t = \nabla \Phi_t = -\nabla F^{\prime} (q_t(\x))$.
\end{proof}
\subsection{Proof of Theorem \ref{th2}}
\begin{proof}
The Lipschitz assumption of $\mathbf{v}_t$ implies the existence and uniqueness of the   McKean-Vlasov equation (15) according to the classical results in ODE \cite{arnold2012geometrical}.
By the uniqueness of the continuity equation, see Proposition 8.1.7 in \cite{ambrosio2008gradient},  it sufficient to show $\mu_t = (\textbf{X}_t)_{\#}\mu$ defined in equation (14) satisfying the continuity equation  (12) in a weak sense. This can be done by the standard test function and soothing approximation arguments, see, Theorem 4.4 in \cite{santambrogio2015optimal} for detail.
\end{proof}
\subsection{Proof of Proposition \ref{prop2}}
\begin{proof}
Without loss of generality let  $K = \frac{T}{s} >1$ be an integer.
Recall $\{\mu_t^{s} \ \  t\in [ks,(k+1)s)$ is  the piecewise linear  interpolation between $\mu_{k}$ and  $\mu_{k+1}$  defined as
$$\mu_t^{s}  = (\mathcal{T}_t^{k,s})_{\#}\mu_k,$$
where, $$\mathcal{T}_t^{k,s}= \textbf{I} + (t-ks) \textbf{v}_{k},$$ $\mu_k$ is defined in (16)-(18) with  $\textbf{v}_{k}= \textbf{v}_{ks}$, i.e., the continuous velocity in (13)  at time $ks$,   $k = 0,.., K-1$, $\mu_0 = \mu.$
  Under the assumption that the velocity fields $\textbf{v}_{t}$ is  Lipschitz continuous on $(\x,\mu_t)$,
    we can first show similarly  as  Lemma 10 in \cite{arbel2019maximum}   $$\mathcal{W}_2(\mu_{ks}, \mu_k) = \mathcal{O}(s).
  \quad \quad  \quad \quad (A1)$$
  Let $\Gamma$ be the optimal coupling between $\mu_k$ and $\mu_{ks}$, and $(X,Y) \sim \Gamma$.
  Let $X_t = \mathcal{T}_t^{k,s}(X)$ and  $Y_t$ be the solution of  (15)  with $\mathbf{X}_0 = Y$ and $t\in [ks,(k+1)s)$.
   Then $$X_t \sim \mu_t^s, \ \  Y_t \sim \mu_t$$ and  $$Y_t = Y + \int_{ks}^{t} \mathbf{v}_{\tilde{t}}(Y_{\tilde{t}}) \mathrm{d} \tilde{t}.$$
      \begin{align*}
   &\mathcal{W}_2^{2}(\mu_t, \mu_{ks})\leq \mathbb{E} [\|Y_t -Y\|_2^2]\\
   &= \mathbb{E} [\|\int_{ks}^{t} \mathbf{v}_{\tilde{t}}(Y_{\tilde{t}}) \mathrm{d} \tilde{t}\|_2^2]\\
   &\leq \mathbb{E}[(\int_{ks}^{t} \|\mathbf{v}_{\tilde{t}}(Y_{\tilde{t}})\|_2\mathrm{d} {\tilde{t}})^2]\\
   &\leq  \mathcal{O}(s^2).   \quad  \quad \quad \quad  (A2)
   \end{align*}
   where, the first inequality follows from the definition of $\mathcal{W}_2$, and the last equality follows from the  the uniform bounded assumption of $\mathbf{v}_t$.
   Similarly,
   \begin{align*}
   &\mathcal{W}_2^{2}(\mu_k, \mu_{t}^{s})\leq \mathbb{E} [\|X -X_t\|_2^2]\\
   &= \mathbb{E} [\|(t-ks)\mathbf{v}_{k}(X)\|_2^2]\\
   &\leq  \mathcal{O}(s^2).   \quad  \quad \quad \quad  (A3)
   \end{align*}
   Then,
 \begin{align*}
  &\mathcal{W}_2(\mu_t, \mu_t^s)\\
  & \leq \mathcal{W}_2(\mu_t, \mu_{ks}) + \mathcal{W}_2(\mu_{ks}, \mu_k)+ \mathcal{W}_2(\mu_{k}, \mu_t^s)\\
  &\leq \mathcal{O}(s),
  \end{align*}
where the first inequality follows from the triangle inequality, see for example Lemma 5.3 in  \cite{santambrogio2015optimal}, and the second  one follows from $(A1)-(A3)$.
\end{proof}
\subsection{Proof of Lemma \ref{lem2}}
\begin{proof}
By  definition,
\begin{equation*}
F(q_t(\x))  =
\left\{\begin{array}{ll}
p(\x) f(\frac{q_t(\x)}{p(\x)}), \ \  \mathcal{L}[\mu] = \mathbb{D}_f(\mu \Vert \nu), \\
(q_t(\x)-p(\x))^2,  \ \ \mathcal{L}[\mu] = \|\mu-\nu\|^2_{L^2(\mathbb{R}^{m})}.
\end{array}
\right.
\end{equation*}
Direct calculation shows
\begin{equation*}
F^{\prime}(q_t(\x))  =
\left\{\begin{array}{ll}
f^{\prime}(\frac{q_t(\x)}{p(\x)}), \ \  \mathcal{L}[\mu] = \mathbb{D}_f(\mu \Vert \nu), \\
2(q_t(\x)-p(\x)),  \ \ \mathcal{L}[\mu] = \|\mu-\nu\|^2_{L^2(\mathbb{R}^{m})}.
\end{array}
\right.
\end{equation*}
Then, the desired  result follows from the above display and  equation (13).
\end{proof}
\subsection{Proof of Lemma \ref{lem3}}
\begin{proof}
By definition, it is easy to check
$$\mathfrak{B}^{0}_{\rm LSDR}(R) = \mathfrak{B}_{\rm ratio}(r, R) - \mathfrak{B}_{\rm ratio}(r, r),$$
where, $$\mathfrak{B}_{\rm ratio}(r, R)$$
the Bregman score with the base probability measure $p$ between $R$ and $r$.
Then $r \in \arg\min_{\text{measureable}\, R } \mathfrak{B}^{0}_{\rm LSDR}(R)$ follow from the fact
$\mathfrak{B}_{\rm ratio}(r, R) \ge \mathfrak{B}_{\rm ratio}(r, r)$ and  the equality holds iff $R = r$.
Since $$\mathfrak{B}^{\alpha}(R) = \mathfrak{B}^{0}_{\rm LSDR}(R) +  \alpha \mathbb{E}_{p} [\Vert \nabla R \Vert_2^2]\geq 0,$$
Then, $$\mathfrak{B}^{\alpha}(R) = 0$$ iff $$\mathfrak{B}^{0}_{\rm LSDR}(R) = 0 \ \ \mathrm{and}  \ \  \mathbb{E}_{p} [\Vert \nabla R \Vert_2^2] = 0,$$ which is further equivalent to
$$R = r = \mathrm{constant}  \ \ (q, p)\text{-}a.e.$$  $\mathrm{constant}  = 1$  due to $r$ is a density ratio.
\end{proof}
\subsection{Proof of Theorem \ref{th3}}
\begin{proof}
We use  $\mathfrak{B}(R)$ to denote $\mathfrak{B}_{\rm LSDR}^{0}-C$ for simplicity, i.e.,
$$\mathfrak{B}(R) = \mathbb{E}_{X \sim p} [ R(X) ^2 ]
- 2 \mathbb{E}_{X \sim q} [R(X)]. \quad (A4) $$
Rewrite  (20) with $\alpha = 0$ as
\begin{align*}
&\widehat{R}_{\phi} \in \arg \min_{R_\phi\in \mathcal{H}_{\mathcal{D}, \mathcal{W}, \mathcal{S}, \mathcal{B}}} \widehat{\mathfrak{B}}(R_{\phi}) \\
&=  \sum_{i=1}^n   \frac{1}{n}(R_{\phi}(X_i)^2
  -2R_{\phi}(Y_i)). \quad (A5)
  \end{align*}
  By Lemma \ref{lem3} and Fermat's rule \cite{clarke1990optimization}, we know $ \mathbf{0} \in \partial  \mathfrak{B}(r).$
  Then,  $\forall R $      direct calculation yields,
  \begin{align*}
  \|R - r\|_{L^2(\nu)}^2  &= \mathfrak{B}(R) - \mathfrak{B}(r) - \langle \partial  \mathfrak{B}(r),  R-r \rangle \\
  &= \mathfrak{B}(R)- \mathfrak{B}(r). \quad (A6)
  \end{align*}
   $ \forall \bar{R}_{\phi} \in \mathcal{H}_{\mathcal{D}, \mathcal{W}, \mathcal{S}, \mathcal{B}} $ we have,
  \begin{align*}
  &\|\widehat{R}_{\phi} - r\|_{L^2(\nu)}^2 = \mathfrak{B}(\widehat{R}_{\phi})- \mathfrak{B}(r) \\
 & = \mathfrak{B}(\widehat{R}_{\phi}) - \widehat{\mathfrak{B}}(\widehat{R}_{\phi}) +    \widehat{\mathfrak{B}}(\widehat{R}_{\phi})-  \widehat{\mathfrak{B}}(\bar{R}_{\phi}) \\
  & +   \widehat{\mathfrak{B}}(\bar{R}_{\phi}) -  \mathfrak{B}(\bar{R}_{\phi}) + \mathfrak{B}(\bar{R}_{\phi})  -  \mathfrak{B}(r)\\
  &\leq  2 \sup_{R \in \mathcal{H}_{\mathcal{D}, \mathcal{W}, \mathcal{S}, \mathcal{B}}} |\mathfrak{B}(R) - \widehat{\mathfrak{B}}(R) |+ \|\bar{R}_{\phi} - r\|_{L^2(\nu)}^2, (A7)
  \end{align*}
  where the  inequality uses the definition of  $\widehat{R}_{\phi}$ and $\bar{R}_{\phi}$ and  (A6).
  We prove our this theorem by upper bounding  the expected value of the right hand side term in (A7). To this end, we need the following
  auxiliary results (A8)-(A10).
  $$ \mathbb{E}_{\{Z_i\}_{i}^n} [\sup_{R } |\mathfrak{B}(R) - \widehat{\mathfrak{B}}(R) | ] \leq 4C_1(2\mathcal{B}+1) \mathfrak{G}(\mathcal{H}), \quad (A8)$$ where $$\mathfrak{G}(\mathcal{H}) =  \mathbb{E}_{\{Z_i, \epsilon_i \}_{i}^n}[\sup_{R\in \mathcal{H}_{\mathcal{D}, \mathcal{W}, \mathcal{S}, \mathcal{B}}}|\frac{1}{n}\sum_{i=1}^n\epsilon_i R(Z_i)|]$$ is the Gaussian complexity of $\mathcal{H}_{\mathcal{D}, \mathcal{W}, \mathcal{S}, \mathcal{B}}$ \cite{bartlett2002rademacher}.\\
  \textbf{Proof of (A8)}.\\
  Let $g(c) = c^2 - c$,  $\mathbf{z} = (\x,\y) \in \mathbb{R}^m \times \mathbb{R}^m$,  $$\widetilde{R}(\mathbf{z}) = (g\circ R)(\mathbf{z}) = R^2(\x) -R(\y).$$
  Denote  $Z = (X,Y)$, $Z_i = (X_i,Y_i), i = 1,...,n$ with $X, X_i$ i.i.d.  $\sim p$, $Y, Y_i$ i.i.d. $\sim q$.
  Let $\widetilde{Z}_i$ be a i.i.d. copy of $Z_i,$ and $\sigma_i (\epsilon_i) $ be the i.i.d. Rademacher random (standard  normal) variables that are independent with
  $Z_i$ and $\widetilde{Z}_i$.
   Then,
   $$\mathfrak{B}(R) = \mathbb{E}_{Z} [\widetilde{R}(Z)] = \frac{1}{n}\mathbb{E}_{\widetilde{Z}_i} [\widetilde{R}(\widetilde{Z}_i)],$$ and
   $$ \widehat{\mathfrak{B}}(R) = \frac{1}{n}\sum_{i=1}^n \widetilde{R}(Z_i).$$
     Denote $$\mathfrak{R}(\mathcal{H}) = \frac{1}{n} \mathbb{E}_{\{Z_i, \sigma_i \}_{i}^n}[\sup_{R\in \mathcal{H}_{\mathcal{D}, \mathcal{W}, \mathcal{S}, \mathcal{B}}}|\sum_{i=1}^n\epsilon_i R(Z_i)|]$$ as the 	
Rademacher complexity of $\mathcal{H}_{\mathcal{D}, \mathcal{W}, \mathcal{S}, \mathcal{B}}$ \cite{bartlett2002rademacher}.
    Then,
  \begin{align*}
  &\mathbb{E}_{\{Z_i\}_{i}^n} [\sup_{R } |\mathfrak{B}(R) - \widehat{\mathfrak{B}}(R) | ] \\
  &=\frac{1}{n} \mathbb{E}_{\{Z_i\}_{i}^n} [\sup_{R } |\sum_{i=1}^n (\mathbb{E}_{\widetilde{Z}_i} [\widetilde{R}(\widetilde{Z}_i)] - \widetilde{R}(Z_i))|]\\
  & \leq  \frac{1}{n} \mathbb{E}_{\{Z_i, \widetilde{Z}_i\}_{i}^n} [\sup_{R } |\widetilde{R}(\widetilde{Z}_i) - \widetilde{R}(Z_i)|]\\
  & =  \frac{1}{n} \mathbb{E}_{\{Z_i, \widetilde{Z}_i,\sigma_i \}_{i}^n} [\sup_{R } |\sum_{i=1}^n\sigma_i(\widetilde{R}(\widetilde{Z}_i) - \widetilde{R}(Z_i))|]\\
  & \leq \frac{1}{n}  \mathbb{E}_{\{Z_i, \sigma_i \}_{i}^n} [\sup_{R } |\sum_{i=1}^n\sigma_i \widetilde{R}(Z_i)| ] \\
  &+ \frac{1}{n}  \mathbb{E}_{\{\widetilde{Z}_i, \sigma_i \}_{i}^n} [\sup_{R } |\sum_{i=1}^n\sigma_i \widetilde{R}(\widetilde{Z}_i)| ] \\
  &= 2\mathfrak{R}(g\circ\mathcal{H})\\
  &\leq 4(2\mathcal{B}+1)\mathfrak{R}(\mathcal{H})\\
  & \leq 4C_1(2\mathcal{B}+1) \mathfrak{G}(\mathcal{H}),
  \end{align*}
  where, the first inequality follows from the Jensen's inequality, and the second equality holds since the distribution of $\sigma_i(\widetilde{R}(\widetilde{Z}_i) - \widetilde{R}(Z_i))$ and $\widetilde{R}(\widetilde{Z}_i) - \widetilde{R}(Z_i)$ are the same, and the last equality holds since the distribution of the two terms are the same, and last two inequality follows from the  Lipschitz contraction property where the Lipschitz constant of $g$ on $\mathcal{H}_{\mathcal{D}, \mathcal{W}, \mathcal{S}, \mathcal{B}}$ is bounded by $2\mathcal{B}+1$ and the relationship between the Gaussian complexity and  the Rademacher complexity, see for Theorem 12 and Lemma 4 in \cite{bartlett2002rademacher}, respectively.
  \begin{align*}
  \mathfrak{G}(\mathcal{H}) \leq 
  C_2\mathcal{B} \sqrt{\frac{n}{\mathcal{D}\mathcal{S}\log \mathcal{S}}}\log \frac{n}{\mathcal{D}\mathcal{S}\log \mathcal{S}} \exp(-\log^2 \frac{n}{\mathcal{D}\mathcal{S}\log \mathcal{S}}).\quad (A9)
  \end{align*}
  \textbf{Proof of (A9)}.\\
  Since $\mathcal{H}$ is negation closed,
\begin{align*}
&\mathfrak{G}(\mathcal{H}) =  \mathbb{E}_{\{Z_i, \epsilon_i \}_{i}^n}[\sup_{R\in \mathcal{H}_{\mathcal{D}, \mathcal{W}, \mathcal{S}, \mathcal{B}}}\frac{1}{n}\sum_{i=1}^n\epsilon_i R(Z_i)]\\
& = \mathbb{E}_{Z_i }[ \mathbb{E}_{\epsilon_i}[\sup_{R\in \mathcal{H}_{\mathcal{D}, \mathcal{W}, \mathcal{S}, \mathcal{B}}}\frac{1}{n}\sum_{i=1}^n\epsilon_i R(Z_i)]|\{Z_i\}_{i=1}^n].
\end{align*}
Conditioning on $\{Z_i\}_{i =1}^n$,
$\forall R, \widetilde{R} \in \mathcal{H}_{\mathcal{D}, \mathcal{W}, \mathcal{S}, \mathcal{B}}$ it easy to check $$\mathbb{V}_{\epsilon_i} [\frac{1}{n}\sum_{i=1}^n\epsilon_i (R(Z_i) - \widetilde{R}(Z_i))] = \frac{d^{\mathcal{H}}_{2}(R,\tilde{R})}{\sqrt{n}},$$
where, $d^{\mathcal{H}}_2(R,\tilde{R}) = \frac{1}{\sqrt{n}} \sqrt{\sum_{i =1}^n (R(Z_i)-\tilde{R}(Z_i))^2}$.
Observing the diameter of $\mathcal{H}_{\mathcal{D}, \mathcal{W}, \mathcal{S}, \mathcal{B}}$ under $d^{\mathcal{H}}_2 $ is at most $\mathcal{B}$,  we have
  \begin{align*}\mathfrak{G}(\mathcal{H}) &\leq \frac{C_3}{\sqrt{n}} \mathbb{E}_{\{Z_i\}_{i=1}^n}[\int_{0}^{B} \sqrt{\log \mathcal{N}(\mathcal{H}, d^{\mathcal{H}}_{2}, \delta)} \mathrm{d} \delta]\\
  &\leq\frac{C_3}{\sqrt{n}} \mathbb{E}_{\{Z_i\}_{i=1}^n}[\int_{0}^{\mathcal{B}} \sqrt{\log \mathcal{N}(\mathcal{H}, d^{\mathcal{H}}_{\infty}, \delta)} \mathrm{d} \delta]\\
  &\leq \frac{C_3}{\sqrt{n}}\int_{0}^{\mathcal{B}} \sqrt{\mathrm{VC}_{\mathcal{H}} \log \frac{6Bn}{\delta \mathrm{VC}_{\mathcal{H}} }} \mathrm{d} \delta, \\
  & \leq C_4 \mathcal{B}(\frac{n}{\mathrm{VC}_{\mathcal{H}}})^{1/2}\log (\frac{n}{\mathrm{VC}_{\mathcal{H}}}) \exp( -\log^2(\frac{n}{\mathrm{VC}_{\mathcal{H}}}))\\
  & \leq C_2\mathcal{B} \sqrt{\frac{n}{\mathcal{D}\mathcal{S}\log \mathcal{S}}}\log \frac{n}{\mathcal{D}\mathcal{S}\log \mathcal{S}} \exp(-\log^2 \frac{n}{\mathcal{D}\mathcal{S}\log \mathcal{S}})
  \end{align*}
  where, the first inequality follows from the chaining  Theorem 8.1.3 in \cite{vershynin2018high}, and the second inequality holds due to
  $d^{\mathcal{H}}_{2}\leq d^{\mathcal{H}}_{\infty}$, and in the third inequality we used
  the relationship between the matric entropy and the VC-dimension of the ReLU networks  $\mathcal{H}_{\mathcal{D}, \mathcal{W}, \mathcal{S}, \mathcal{B}}$ \cite{anthony2009neural}, i.e.,
  $$\log \mathcal{N}(\mathcal{H}, d^{\mathcal{H}}_{\infty}, \delta) \leq \mathrm{VC}_{\mathcal{H}} \log \frac{6\mathcal{B}n}{\delta\mathrm{VC}_{\mathcal{H}}},$$
  and the fourth inequality follows by  some calculation,
  and the last inequality  holds due to the  upper bound of VC-dimension for the ReLU network $\mathcal{H}_{\mathcal{D}, \mathcal{W}, \mathcal{S}, \mathcal{B}}$ satisfying  $$\mathrm{VC}_{\mathcal{H}} \leq C_5 \mathcal{D}\mathcal{S}\log \mathcal{S},$$ see \cite{bartlett2019}.\\
  For any two integer $M,N$, there exists  a  $\bar{R}_{\phi} \in \mathcal{H}_{\mathcal{D}, \mathcal{W}, \mathcal{S}, \mathcal{B}}$  with width
 $$\mathcal{W} = \max\{8 \mathcal{M} N^{1 /\mathcal{M}} +4 \mathcal{M}, 12 N+14\},$$ and
 depth  $$\mathcal{D} = 9M+12,$$ and $\mathcal{B} = 2B,$
 such that
  \begin{align*}
  \|r-\bar{R}_{\phi}\|^2_{L^{2}(\nu)}  \leq C_6    L m \mathcal{M} (NM)^{-4/\mathcal{M}}. \quad (A10).
  \end{align*}
    \textbf{Proof of (A10)}.\\
    We use Lemma 4.1, Theorem 4.3, 4.4   and following  the proof of Theorem 1.3 in  \cite{shen2019deep}.
    Let $\mathbf{A}$ be the random orthoprojector in Theorem 4.4, then
  it is to check $\mathbf{A}(\mathfrak{M}_{\epsilon}) \subset \mathbf{A}([-c,c]^{m}) \subset [-c\sqrt{m},\sqrt{m}c]^{\mathcal{M}}.$
  Let $\tilde{r}$ be a extension of the restriction of  $r$  on  $\mathfrak{M}_{\epsilon}$, which is defined  similarly  as $\tilde{g}$ on page 30  in  \cite{shen2019deep}.
   Since we assume the target $r$ is Lipschitz continuous with the bound $B$ and the Lipschitz constant $L$, let $\epsilon$ small enough,  then by Theorem 4.3, there exist a ReLU network  $\tilde{R}_{\phi} \in \mathcal{H}_{\mathcal{D}, \mathcal{W}, \mathcal{S}, \mathcal{B}}$  with width
 $$\mathcal{W} = \max\{8 \mathcal{M} N^{1 /\mathcal{M}} +4 \mathcal{M}, 12 N+14\},$$ and
 depth  $$\mathcal{D} = 9M+12,$$ and $\mathcal{B} = 2B,$
 such that
 $$\|\tilde{r} - \tilde{R}_{\phi}\|_{L^{\infty}(\mathfrak{M}_{\epsilon} \setminus \mathcal{N})} \leq 80c L \sqrt{m \mathcal{M}} (NM)^{-2/m},$$
 and
 $$\|\tilde{R}_{\phi}\|_{L^{\infty}(\mathfrak{M}_{\epsilon})} \leq B+3cL \sqrt{m\mathcal{M}},$$
 where, $\mathcal{N}$ is a $\nu-$ negligible set with $\nu(\mathcal{N})$ can be arbitrary small.
 Define $\bar{R}_{\phi} = \tilde{R}_{\phi}\circ \mathbf{A}$. Then, following
 the proof after equation (4.8) in  Theorem 1.3  \cite{shen2019deep}, we get our  (A10) and $$\|\bar{R}_{\phi}\|_{L^{\infty}(\mathfrak{M}_{\epsilon} \setminus \mathcal{N})} \leq 2B, \ \  \|\bar{R}_{\phi}\|_{L^{\infty}(\mathcal{N})} \leq 2B+3cL \sqrt{m\mathcal{M}}.$$
Let $\mathcal{D}\mathcal{S}\log \mathcal{S}< n$, combing the results  $A(7) - A(10)$,
 we have
 \begin{align*}
&\mathbb{E}_{\{X_i,Y_i\}_{1}^n} [\|\widehat{R}_{\phi} - r\|_{L^2(\nu)}^2] \\
&\leq 8C_1(2B+1) \mathfrak{G}(\mathcal{H}) +  C_6   c L m \mathcal{M} (NM)^{-4/\mathcal{M}}\\
&\leq  8C_1(2B+1)C_2 B\sqrt{\frac{\mathcal{D}\mathcal{S}\log \mathcal{S}}{n}}\log \frac{n}{\mathcal{D}\mathcal{S}\log \mathcal{S}}+ C_6    cL m \mathcal{M} (NM)^{-4/\mathcal{M}}\\
& \leq  C(B^2+ cL m \mathcal{M}) n^{-2/(2+ \mathcal{M})},
 \end{align*}
 where, last inequality holds since  we  choose $$M = \log n,$$ $$ N = n^{\frac{\mathcal{M}}{2(2+\mathcal{M})}}/\log n$$,  $$\mathcal{S} = n^{\frac{\mathcal{M}-2}{\mathcal{M}+2}}/\log^4 n,$$ i.e.,
 $$\mathcal{D} = 9 \log n + 12$$, $$\mathcal{W} = 12n^{\frac{\mathcal{M}}{2(2+\mathcal{M})}}/\log n+14.$$
\end{proof}
\subsection{Proof of the relation between UnifiGem  and MMD flow}\label{pstein}
\begin{proof}
Let $\mathcal{H}$ be a reproducing kernel Hilbert space with  characteristic kernel
 $K(\x,\mathbf{z})$.
 Recall in MMD flow, $$\mathcal{L}[\mu] = \frac{1}{2}\|\mu-\nu\|_{\mathrm{mmd}}^2,$$ and
$$\frac{\partial \mathcal{L}[\mu]}{\partial \mu} (\x) = \int K(\x,\mathbf{z}) \mathrm{d}\mu(\mathbf{z}) -\int K(\x,\mathbf{z}) \mathrm{d}\nu(\mathbf{z}),$$
and the vector fields
\begin{align*}
&\mathbf{v}_{t}^{\mathrm{mmd}} = -\nabla \frac{\partial \mathcal{L}[\mu]}{\partial \mu_t}\\
 &= \int \nabla_{\x} K(\x,\mathbf{z}) \mathrm{d}\nu(\mathbf{z})-\int \nabla_{\x} K(\x,\mathbf{z}) \mathrm{d}\mu_t(\mathbf{z})\\
 & =  \int \nabla_{\x} K(\x,\mathbf{z})  p(\mathbf{z}) \mathrm{d} \mathbf{z}-\int \nabla_{\x} K(\x,\mathbf{z})q_t(\mathbf{z}) \mathrm{d}\mathbf{z}
\end{align*}
By Lemma \ref{lem2}, the vector fields corresponding the Lebesgue norm
$\frac{1}{2}\|\mu-\nu\|^2_{L^2(\mathbb{R}^{m})} =  \frac{1}{2}\int_{\mathbb{R}^{m}} |q(\x)- p(\x)|^2  {\mathrm{d}} \x$  are defined as
$$\mathbf{v}_{t} =  \nabla p(\x) - \nabla q_t(\x).$$
Next, we will show the vector fields $\mathbf{v}_{t}^{\mathrm{mmd}}$ is exactly by projecting  the vector fields $\mathbf{v}_{t}$ on to
the reproducing kernel Hilbert space $\mathcal{H}^{m} = \mathcal{H}^{\otimes m} $.
By the definition of reproducing kernel we have,
$$p(\x) = \left \langle p(\cdot), K(\x,\cdot) \right \rangle_{\mathcal{H}} = \int K(\x,\mathbf{z}) p(\mathbf{z})\mathrm{d}\mathbf{z},$$ and $$q_t(\x) = \left \langle q_t(\cdot), K(\x,\cdot) \right \rangle_{\mathcal{H}} = \int K(\x,\mathbf{z}) q_t(\mathbf{z})\mathrm{d}\mathbf{z}.$$
Hence,
\begin{align*}
&\mathbf{v}_{t}(\x ) = \nabla p(\x) - \nabla q_t(\x) \\
=&\int \nabla_{x} K(\x,\mathbf{z})( p(\mathbf{z}) -  q_t(\mathbf{z}))\mathrm{d}\mathbf{z}\\
=& \mathbf{v}_{t}^{\mathrm{mmd}}(\x).
\end{align*}
\end{proof}

\bibliographystyle{abbrv}

\begin{thebibliography}{10}

\bibitem{ali1966general}
S.~M. Ali and S.~D. Silvey.
\newblock A general class of coefficients of divergence of one distribution
  from another.
\newblock {\em Journal of the Royal Statistical Society: Series B
  (Methodological)}, 28(1):131--142, 1966.

\bibitem{ambrosio2008gradient}
L.~Ambrosio, N.~Gigli, and G.~Savar{\'e}.
\newblock {\em Gradient flows: in metric spaces and in the space of probability
  measures}.
\newblock Springer Science \& Business Media, 2008.

\bibitem{arbel2019maximum}
M.~Arbel, A.~Korba, A.~Salim, and A.~Gretton.
\newblock Maximum mean discrepancy gradient flow.
\newblock In {\em NeurIPS}, 2019.

\bibitem{arbel18}
M.~Arbel, D.~Sutherland, M.~Bi\'{n}kowski, and A.~Gretton.
\newblock On gradient regularizers for {MMD} {GAN}s.
\newblock In {\em NeurIPS}, 2018.

\bibitem{arjovsky2017principled}
M.~Arjovsky and L.~Bottou.
\newblock Towards principled methods for training generative adversarial
  networks.
\newblock In {\em ICLR}, 2017.

\bibitem{arjovsky17}
M.~Arjovsky, S.~Chintala, and L.~Bottou.
\newblock Wasserstein generative adversarial networks.
\newblock In {\em ICML}, 2017.

\bibitem{bauer2019deep}
B.~Bauer, M.~Kohler, et~al.
\newblock On deep learning as a remedy for the curse of dimensionality in
  nonparametric regression.
\newblock {\em The Annals of Statistics}, 47(4):2261--2285, 2019.

\bibitem{benamou2000computational}
J.-D. Benamou and Y.~Brenier.
\newblock A computational fluid mechanics solution to the monge-kantorovich
  mass transfer problem.
\newblock {\em Numerische Mathematik}, 84(3):375--393, 2000.

\bibitem{binkowski18}
M.~Bi{\'n}kowski, D.~J. Sutherland, M.~Arbel, and A.~Gretton.
\newblock Demystifying {MMD} {GAN}s.
\newblock In {\em ICLR}, 2018.

\bibitem{brenier1991polar}
Y.~Brenier.
\newblock Polar factorization and monotone rearrangement of vector-valued
  functions.
\newblock {\em Communications on pure and applied mathematics}, 44(4):375--417,
  1991.

\bibitem{chen2018continuous}
C.~Chen, C.~Li, L.~Chen, W.~Wang, Y.~Pu, and L.~C. Duke.
\newblock Continuous-time flows for efficient inference and density estimation.
\newblock In {\em ICML}, 2018.

\bibitem{chen2018neural}
T.~Q. Chen, Y.~Rubanova, J.~Bettencourt, and D.~K. Duvenaud.
\newblock Neural ordinary differential equations.
\newblock In {\em NIPS}, 2018.

\bibitem{dawid2007geometry}
A.~P. Dawid.
\newblock The geometry of proper scoring rules.
\newblock {\em Annals of the Institute of Statistical Mathematics},
  59(1):77--93, 2007.

\bibitem{de1993new}
E.~De~Giorgi.
\newblock New problems on minimizing movements. in boundary value problems for
  partial differential equations, res. notes appl. math. vol. 29.
\newblock pages 81--98, 1993.

\bibitem{deshpande2018generative}
I.~Deshpande, Z.~Zhang, and A.~G. Schwing.
\newblock Generative modeling using the sliced wasserstein distance.
\newblock In {\em CVPR}, 2018.

\bibitem{dinh2014nice}
L.~Dinh, D.~Krueger, and Y.~Bengio.
\newblock {NICE}: Non-linear independent components estimation.
\newblock In {\em ICLR}, 2015.

\bibitem{dinh2016density}
L.~Dinh, J.~Sohl-Dickstein, and S.~Bengio.
\newblock Density estimation using {R}eal {NVP}.
\newblock In {\em ICLR}, 2017.

\bibitem{gao2019deep}
Y.~Gao, Y.~Jiao, Y.~Wang, Y.~Wang, C.~Yang, and S.~Zhang.
\newblock Deep generative learning via variational gradient flow.
\newblock In {\em ICML}, 2019.

\bibitem{genevay2018learning}
A.~Genevay, G.~Peyre, and M.~Cuturi.
\newblock Learning generative models with sinkhorn divergences.
\newblock In {\em ICML}, 2018.

\bibitem{gneiting2007strictly}
T.~Gneiting and A.~E. Raftery.
\newblock Strictly proper scoring rules, prediction, and estimation.
\newblock {\em Journal of the American statistical Association},
  102(477):359--378, 2007.

\bibitem{goodfellow14}
I.~Goodfellow, J.~Pouget-Abadie, M.~Mirza, B.~Xu, D.~Warde-Farley, S.~Ozair,
  A.~Courville, and Y.~Bengio.
\newblock Generative adversarial nets.
\newblock In {\em NIPS}, 2014.

\bibitem{grathwohl2019scalable}
W.~Grathwohl, R.~Chen, J.~Bettencourt, and D.~Duvenaud.
\newblock Scalable reversible generative models with free-form continuous
  dynamics.
\newblock In {\em ICLR Workshop}, 2019.

\bibitem{grathwohl2018ffjord}
W.~Grathwohl, R.~T. Chen, J.~Bettencourt, I.~Sutskever, and D.~Duvenaud.
\newblock Ffjord: Free-form continuous dynamics for scalable reversible
  generative models.
\newblock In {\em ICLR}, 2019.

\bibitem{heusel17}
M.~Heusel, H.~Ramsauer, T.~Unterthiner, B.~Nessler, and S.~Hochreiter.
\newblock {GAN}s trained by a two time-scale update rule converge to a local
  nash equilibrium.
\newblock In {\em NIPS}, 2017.

\bibitem{higgins16}
I.~Higgins, L.~Matthey, A.~Pal, C.~Burgess, X.~Glorot, M.~Botvinick,
  S.~Mohamed, and A.~Lerchner.
\newblock $\beta$-{VAE}: Learning basic visual concepts with a constrained
  variational framework.
\newblock In {\em ICLR}, 2017.

\bibitem{johnson18}
R.~Johnson and T.~Zhang.
\newblock Composite functional gradient learning of generative adversarial
  models.
\newblock In {\em ICML}, 2018.

\bibitem{jordan1998variational}
R.~Jordan, D.~Kinderlehrer, and F.~Otto.
\newblock The variational formulation of the fokker--planck equation.
\newblock {\em SIAM journal on mathematical analysis}, 29(1):1--17, 1998.

\bibitem{kanamori2014statistical}
T.~Kanamori and M.~Sugiyama.
\newblock Statistical analysis of distance estimators with density differences
  and density ratios.
\newblock {\em Entropy}, 16(2):921--942, 2014.

\bibitem{kingma2018glow}
D.~P. Kingma and P.~Dhariwal.
\newblock Glow: Generative flow with invertible 1x1 convolutions.
\newblock In {\em NeurIPS}, 2018.

\bibitem{kingma2016improved}
D.~P. Kingma, T.~Salimans, R.~Jozefowicz, X.~Chen, I.~Sutskever, and
  M.~Welling.
\newblock Improved variational inference with inverse autoregressive flow.
\newblock In {\em NIPS}, 2016.

\bibitem{kingma14}
D.~P. Kingma and M.~Welling.
\newblock Auto-encoding variational bayes.
\newblock In {\em ICLR}, 2014.

\bibitem{modelsliced}
S.~Kolouri, P.~E. Pope, C.~E. Martin, and G.~K. Rohde.
\newblock Sliced-wasserstein autoencoder: An embarrassingly simple generative
  model.
\newblock In {\em ICLR}, 2019.

\bibitem{krizhevsky09}
A.~Krizhevsky and G.~Hinton.
\newblock Learning multiple layers of features from tiny images.
\newblock Technical report, Citeseer, 2009.

\bibitem{lecun98}
Y.~LeCun, L.~Bottou, Y.~Bengio, and P.~Haffner.
\newblock Gradient-based learning applied to document recognition.
\newblock {\em Proceedings of the IEEE}, 86(11):2278--2324, 1998.

\bibitem{leveque2007finite}
R.~J. LeVeque.
\newblock {\em Finite difference methods for ordinary and partial differential
  equations: steady-state and time-dependent problems}, volume~98.
\newblock 2007.

\bibitem{li17}
C.-L. Li, W.-C. Chang, Y.~Cheng, Y.~Yang, and B.~P{\'o}czos.
\newblock {MMD GAN}: Towards deeper understanding of moment matching network.
\newblock In {\em NIPS}, 2017.

\bibitem{li15}
Y.~Li, K.~Swersky, and R.~Zemel.
\newblock Generative moment matching networks.
\newblock In {\em ICML}, 2015.

\bibitem{liu2018two}
H.~Liu, G.~Xianfeng, and D.~Samaras.
\newblock A two-step computation of the exact gan wasserstein distance.
\newblock In {\em ICML}, 2018.

\bibitem{liu15}
Z.~Liu, P.~Luo, X.~Wang, and X.~Tang.
\newblock Deep learning face attributes in the wild.
\newblock In {\em ICCV}, 2015.

\bibitem{liutkus2019sliced}
A.~Liutkus, U.~Simsekli, S.~Majewski, A.~Durmus, F.-R. St{\"o}ter,
  K.~Chaudhuri, and R.~Salakhutdinov.
\newblock Sliced-wasserstein flows: Nonparametric generative modeling via
  optimal transport and diffusions.
\newblock In {\em ICML}, 2019.

\bibitem{makhzani15}
A.~Makhzani, J.~Shlens, N.~Jaitly, I.~Goodfellow, and B.~Frey.
\newblock Adversarial autoencoders.
\newblock In {\em ICLR}, 2016.

\bibitem{mao17}
X.~Mao, Q.~Li, H.~Xie, R.~Y. Lau, Z.~Wang, and S.~P. Smolley.
\newblock Least squares generative adversarial networks.
\newblock In {\em ICCV}, 2017.

\bibitem{mccann1995existence}
R.~J. McCann et~al.
\newblock Existence and uniqueness of monotone measure-preserving maps.
\newblock {\em Duke Mathematical Journal}, 80(2):309--324, 1995.

\bibitem{mohamed2016learning}
S.~Mohamed and B.~Lakshminarayanan.
\newblock Learning in implicit generative models.
\newblock {\em arXiv preprint arXiv:1610.03483}, 2016.

\bibitem{mroueh17}
Y.~Mroueh and T.~Sercu.
\newblock {F}isher {GAN}.
\newblock In {\em NIPS}, 2017.

\bibitem{nowozin16}
S.~Nowozin, B.~Cseke, and R.~Tomioka.
\newblock $f$-{GAN}: Training generative neural samplers using variational
  divergence minimization.
\newblock In {\em NIPS}, 2016.

\bibitem{papamakarios2017masked}
G.~Papamakarios, T.~Pavlakou, and I.~Murray.
\newblock Masked autoregressive flow for density estimation.
\newblock In {\em NIPS}, 2017.

\bibitem{patrinisinkhorn}
G.~Patrini, S.~Bhargav, R.~van~den Berg, M.~Welling, P.~Forr{\'e}, T.~Genewein,
  M.~Carioni, K.~Graz, F.~Nielsen, and C.~Sony.
\newblock Sinkhorn autoencoders.
\newblock In {\em UAI}, 2019.

\bibitem{reed16}
S.~Reed, Z.~Akata, X.~Yan, L.~Logeswaran, B.~Schiele, and H.~Lee.
\newblock Generative adversarial text to image synthesis.
\newblock In {\em ICML}, 2016.

\bibitem{rezende2015variational}
D.~J. Rezende and S.~Mohamed.
\newblock Variational inference with normalizing flows.
\newblock In {\em ICML}, 2015.

\bibitem{roth2017stabilizing}
K.~Roth, A.~Lucchi, S.~Nowozin, and T.~Hofmann.
\newblock Stabilizing training of generative adversarial networks through
  regularization.
\newblock In {\em NIPS}, pages 2018--2028, 2017.

\bibitem{salakhutdinov2015learning}
R.~Salakhutdinov.
\newblock Learning deep generative models.
\newblock {\em Annual Review of Statistics and Its Application}, 2:361--385,
  2015.

\bibitem{santambrogio2015optimal}
F.~Santambrogio.
\newblock {\em Optimal transport for applied mathematicians}.
\newblock Springer, 2015.

\bibitem{sonderby2016amortised}
C.~K. S{\o}nderby, J.~Caballero, L.~Theis, W.~Shi, and F.~Husz{\'a}r.
\newblock Amortised map inference for image super-resolution.
\newblock In {\em ICLR}, 2017.

\bibitem{srip12}
B.~K. Sriperumbudur, K.~Fukumizu, A.~Gretton, B.~Sch{\"o}lkopf, G.~R.
  Lanckriet, et~al.
\newblock On the empirical estimation of integral probability metrics.
\newblock {\em Electronic Journal of Statistics}, 6:1550--1599, 2012.

\bibitem{sugiyama2012density2}
M.~Sugiyama, T.~Kanamori, T.~Suzuki, M.~D. Plessis, S.~Liu, and I.~Takeuchi.
\newblock Density-difference estimation.
\newblock In {\em NIPS}, 2012.

\bibitem{sugiyama2012density}
M.~Sugiyama, T.~Suzuki, and T.~Kanamori.
\newblock {\em Density ratio estimation in machine learning}.
\newblock Cambridge University Press, 2012.

\bibitem{sutherland16}
D.~J. Sutherland, H.-Y. Tung, H.~Strathmann, S.~De, A.~Ramdas, A.~Smola, and
  A.~Gretton.
\newblock Generative models and model criticism via optimized maximum mean
  discrepancy.
\newblock In {\em ICLR}, 2017.

\bibitem{tao18}
C.~Tao, L.~Chen, R.~Henao, J.~Feng, and L.~C. Duke.
\newblock {C}hi-square generative adversarial network.
\newblock In {\em ICML}, 2018.

\bibitem{tolstikhin18}
I.~Tolstikhin, O.~Bousquet, S.~Gelly, and B.~Schoelkopf.
\newblock Wasserstein auto-encoders.
\newblock In {\em ICML}, 2018.

\bibitem{villani2008optimal}
C.~Villani.
\newblock {\em Optimal transport: old and new}, volume 338.
\newblock Springer Science \& Business Media, 2008.

\bibitem{zhang2018monge}
L.~Zhang, L.~Wang, et~al.
\newblock Monge-ampere flow for generative modeling.
\newblock {\em arXiv preprint arXiv:1809.10188}, 2018.

\bibitem{zhang2019wasserstein}
S.~Zhang, Y.~Gao, Y.~Jiao, J.~Liu, Y.~Wang, and C.~Yang.
\newblock Wasserstein-wasserstein auto-encoders.
\newblock {\em arXiv preprint arXiv:1902.09323}, 2019.

\bibitem{zhu17}
J.-Y. Zhu, T.~Park, P.~Isola, and A.~A. Efros.
\newblock Unpaired image-to-image translation using cycle-consistent
  adversarial networks.
\newblock In {\em ICCV}, 2017.



\bibitem[Anthony \& Bartlett(2009)]{anthony2009neural}
Anthony, M. and Bartlett, P.~L.
\newblock \emph{Neural network learning: Theoretical foundations}.
\newblock cambridge university press, 2009.


\bibitem[Arnold(2012)]{arnold2012geometrical}
Arnold, V.~I.
\newblock \emph{Geometrical methods in the theory of ordinary differential
  equations}, volume 250.
\newblock Springer Science \& Business Media, 2012.

\bibitem[Bartlett \& Mendelson(2002)]{bartlett2002rademacher}
Bartlett, P.~L. and Mendelson, S.
\newblock Rademacher and gaussian complexities: Risk bounds and structural
  results.
\newblock \emph{Journal of Machine Learning Research}, 3:\penalty0 463--482,
  2002.

\bibitem[Bartlett et~al.(2019)]{bartlett2019}
Bartlett, P.~L., Harvey, N., Liaw, C., and Mehrabian, A.
\newblock Nearly-tight vc-dimension and pseudodimension bounds for piecewise
  linear neural networks.
\newblock \emph{Journal of Machine Learning Research}, 20:\penalty0 1--17,
  2019.


\bibitem[Clarke(1990)]{clarke1990optimization}
Clarke, F.~H.
\newblock \emph{Optimization and nonsmooth analysis}, volume~5.
\newblock Siam, 1990.

\bibitem[Gelfand \& Silverman(2000)]{gelfand2000calculus}
Gelfand, I.~M., Silverman, R.~A., et~al.
\newblock \emph{Calculus of variations}.
\newblock 2000.





\bibitem[Shen et~al.(2019)]{shen2019deep}
Shen, Z., Yang, H., and Zhang, S.
\newblock Deep network approximation characterized by number of neurons.
\newblock \emph{arXiv preprint arXiv:1906.05497}, 2019.

\bibitem[Vershynin(2018)]{vershynin2018high}
Vershynin, R.
\newblock \emph{High-dimensional probability: An introduction with applications
  in data science}, volume~47.
\newblock Cambridge university press, 2018.


\bibitem[Heusel et~al.(2017)]{heusel17}
Heusel, M., Ramsauer, H., Unterthiner, T., Nessler, B., and Hochreiter, S.
\newblock {GAN}s trained by a two time-scale update rule converge to a local
  nash equilibrium.
\newblock In \emph{NIPS}, 2017.

\bibitem[Ioffe \& Szegedy(2015)]{ioffe2015batch}
Ioffe, S. and Szegedy, C.
\newblock Batch normalization: Accelerating deep network training by reducing
  internal covariate shift.
\newblock In \emph{ICML}, 2015.



\bibitem[Miyato et~al.(2018)]{miyato18}
Miyato, T., Kataoka, T., Koyama, M., and Yoshida, Y.
\newblock Spectral normalization for generative adversarial networks.
\newblock In \emph{ICLR}, 2018.


\bibitem[Szegedy et~al.(2016)]{szegedy16}
Szegedy, C., Vanhoucke, V., Ioffe, S., Shlens, J., and Wojna, Z.
\newblock Rethinking the inception architecture for computer vision.
\newblock In \emph{CVPR}, 2016.

\bibitem[Chertock(2017)]{chertock2017practical}
Chertock, A.
\newblock A practical guide to deterministic particle methods.
\newblock In \emph{Handbook of numerical analysis}, volume~18, pp.\  177--202.
  Elsevier, 2017.


\end{thebibliography}


\end{document}